\documentclass{article} %
\usepackage{iclr2027_conference,times}

\usepackage{amsmath,amsfonts,bm}

\def\eqref#1{equation~\ref{#1}}

\def\1{\bm{1}}

\DeclareMathAlphabet{\mathsfit}{\encodingdefault}{\sfdefault}{m}{sl}
\SetMathAlphabet{\mathsfit}{bold}{\encodingdefault}{\sfdefault}{bx}{n}

\usepackage{hyperref}
\usepackage{url}

\usepackage{graphicx}
\usepackage{booktabs}     %
\usepackage{multirow}
\usepackage{subcaption}
\usepackage{xcolor}
\usepackage{amssymb}
\usepackage{wrapfig}      %
\usepackage{lipsum}
\usepackage[none]{hyphenat}
\usepackage{booktabs}
\usepackage[table]{xcolor}   %
\usepackage{microtype}
\usepackage{threeparttable}
\usepackage{enumitem}
\usepackage{placeins}

\definecolor{hlgreen}{HTML}{4CAF50}   %

\newcommand{\ca}[1]{\cellcolor{hlgreen!12}#1}
\newcommand{\cb}[1]{\cellcolor{hlgreen!6}#1}
\newcommand{\cc}[1]{\cellcolor{hlgreen!3}#1}

\title{Same Scene, Different Task: Skill Alignment for Compositional Generalization in VLAs}

\author{Taegeun Yang, Youngju Na, Yoonki Cho, and Sung-Eui Yoon\\
Korea Advanced Institute of Science and Technology (KAIST)\\
\texttt{\{xorms2001, yjna2907, yoonki, sungeui\}@kaist.ac.kr}
}

\iclrfinalcopy %
\begin{document}

\maketitle

\vspace{-0.3cm}
\begin{figure}[h]
\centering
\captionsetup{font={stretch=1.0}, skip=4pt}
\includegraphics[width=\linewidth]{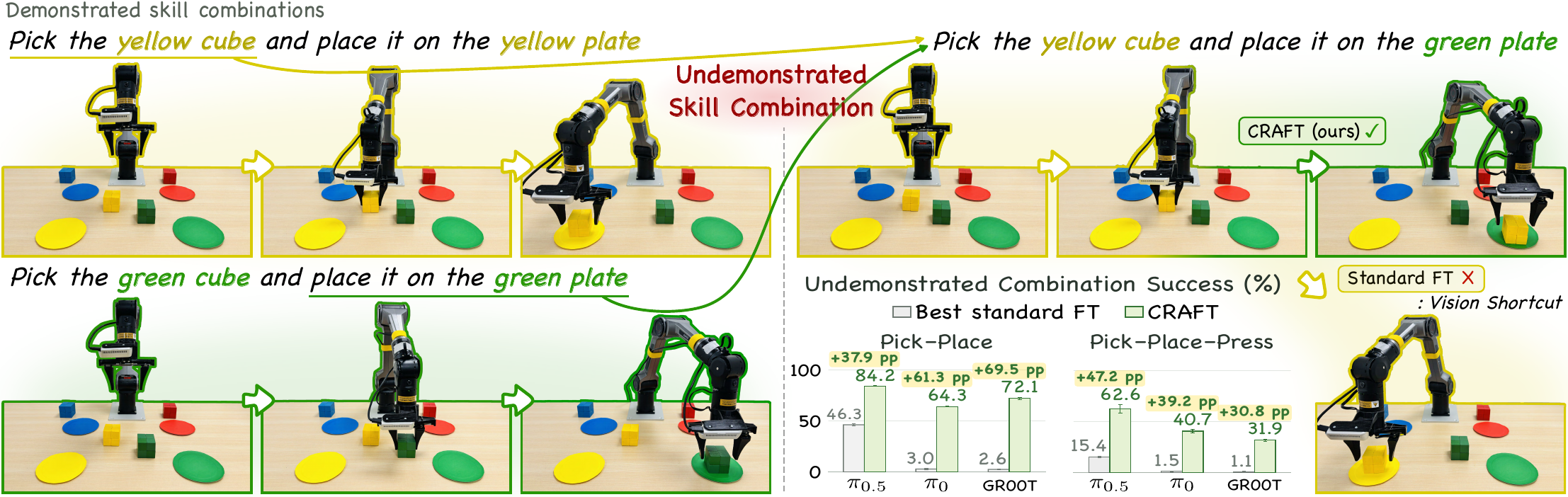}
\caption{\fontsize{10}{10.5}\selectfont
Generalization to undemonstrated skill combinations.
Fine-tuned VLAs struggle to execute new combinations of demonstrated skills.
CRAFT uses demonstrated skill executions to train the policy to follow instructions for new combinations, improving success on undemonstrated combinations across two benchmarks and three VLA models over standard fine-tuning.}
\label{fig:teaser}
\end{figure}

\begin{abstract}
\vspace{-0.18cm}
Vision-language-action (VLA) models often struggle to generalize to skill combinations absent from their fine-tuning demonstrations, even when every constituent skill has been demonstrated.
We focus on a \emph{vision shortcut} as one failure mode: during fine-tuning, visual observations can serve as a proxy for the instruction, so a policy may execute a demonstrated combination associated with similar observations rather than the instructed combination.
This motivates training with counterfactual pairs formed by holding a demonstration observation fixed while changing the instruction to specify an undemonstrated combination.
These pairs, however, lack corresponding demonstrated action targets.
Crucially, the currently required skill has already been demonstrated, but actions from those executions cannot serve as direct targets because the same skill can require different actions across observations.
We propose CRAFT, which transfers supervision from demonstrated executions of the required skill to counterfactual pairs using skill representations that can be reused across executions of the same skill.
Across three VLA models and two simulation benchmarks, CRAFT improves success on undemonstrated combinations while maintaining high success on demonstrated ones; it also improves compositional generalization on a real robot.
Project \href{https://taegeunyang.github.io/craft/}{\textit{Website}}.
\end{abstract}

\vspace{-0.2cm}
\section{Introduction}
\label{sec:intro}
\vspace{-0.1cm}
Vision-language-action (VLA) models map visual observations and natural-language instructions to robot actions~\citep{brohan2023rt2, kim2024openvla, black2024pi0, black2025pi05, bjorck2025gr00t}.
Adapting these models to a target robot embodiment, sensor configuration, or environment often involves fine-tuning on robot demonstrations collected in that setting~\citep{kim2025oft}.
Because collecting demonstrations that cover every task of interest is costly, scalable adaptation requires generalization beyond the demonstrated tasks.
Prior work has studied how adapted VLAs retain capabilities from pretraining~\citep{yadav2026robust, guo2026priorvla}, generalize across objects and visual conditions~\citep{dey2025revla}, and remain robust to language variation~\citep{grover2025enhancing}.
In this work, we study a complementary form of generalization: whether a VLA can \emph{recombine skills demonstrated during adaptation into combinations absent from those demonstrations}.

Many robotic tasks can be viewed as compositions of skills, such as picking an object and then placing it at a target location~\citep{ahn2022can, shi2025hi}.
We define a \emph{skill} as an operation paired with the entity specified for that operation, such as picking a red cube or placing an object on a blue plate.
For example, demonstrations of ``\emph{pick the yellow cube and place it on the yellow plate}'' and ``\emph{pick the green cube and place it on the green plate}'' contain all the constituent skills needed for ``\emph{pick the yellow cube and place it on the green plate},'' even though that combination is absent from the demonstrations (Figure~\ref{fig:teaser}).
As the number of possible entities grows, collecting demonstrations for every skill combination becomes increasingly costly.
This motivates a compositional generalization setting in which all required skills appear in the fine-tuning demonstrations, but their combination does not.
Yet even when every constituent skill is demonstrated, we observe that fine-tuned VLAs often perform well on demonstrated combinations but struggle with new combinations (Appendix~\ref{app:compositional-failure}).

\looseness=-1
To address this gap, we focus on one failure mode: a \emph{vision shortcut}, in which fine-tuned VLAs use visual observations as a proxy for the instruction~\citep{xu2025bayesvla, lian2026langforce}. 
Under such a shortcut, a policy may execute a demonstrated combination associated with similar observations even when instructed to perform a new one (Figure~\ref{fig:teaser}). 
To mitigate this shortcut, we form \emph{counterfactual pairs} during fine-tuning by keeping a demonstration observation fixed while changing the instruction to specify a combination absent from the demonstrations.
The resulting pair, however, has no corresponding demonstrated action target.
Crucially, the skill currently required under the changed instruction has already been demonstrated.
Nonetheless, actions from those demonstrations cannot generally serve as direct targets for the counterfactual pair, because the same skill can require different actions across observations.
Thus, the key challenge is to \emph{transfer} supervision from demonstrated executions of the required skill to the counterfactual pair while accounting for how the required action varies across observations.

We propose \textbf{CRAFT} (\textbf{C}ounterfactual Skill-\textbf{R}epresentation \textbf{A}lignment for \textbf{F}ine-\textbf{T}uning), which addresses this challenge by learning skill representations that can be reused across executions of the same skill while allowing action prediction to vary with the observation.
CRAFT introduces a \emph{skill representation} that reflects the currently required skill and a \emph{state representation} derived from the visual observation; together, they condition action prediction.
For each counterfactual pair, CRAFT selects a demonstrated execution of the required skill.
It combines the pair's skill representation with that execution's state representation and supervises the prediction with that execution's action target.
This supervision trains the counterfactual pair's skill representation to reflect the required skill in action prediction.
Thus, CRAFT transfers supervision from demonstrated executions of the required skill to the counterfactual pair without requiring a demonstrated action target for that pair.

We evaluate CRAFT across $\pi$-family~\citep{black2024pi0, black2025pi05} and GR00T~\citep{bjorck2025gr00t} VLA models on two compositional benchmarks in simulation, with additional evaluation of $\pi_{0.5}$ on a real robot.
CRAFT substantially improves success on undemonstrated skill combinations while maintaining high performance on demonstrated combinations.
Our contributions are as follows:
\begin{itemize}[leftmargin=1.2em, topsep=0pt, itemsep=2pt, parsep=0.5pt]
    \item We propose \textbf{CRAFT}, a counterfactual supervision approach that uses existing demonstrations of constituent skills to train VLAs to execute skill combinations absent from the original demonstrations, without collecting demonstrations for those combinations.

    \item We introduce two compositional benchmarks for two- and three-skill tasks, with demonstration datasets that cover every constituent skill but only a subset of skill combinations; we will release the datasets and simulation environments.

    \item Across $\pi$-family and GR00T VLA models, CRAFT improves success on skill combinations absent from the original demonstrations while maintaining high success on demonstrated combinations; we additionally validate CRAFT with $\pi_{0.5}$ on a real robot.
\end{itemize}

\section{Related Work}
\label{sec:related}

\subsection{Compositional Generalization in VLAs}
\label{sec:related-vla}
Compositional generalization concerns generalizing to new combinations of components encountered during training~\citep{lake2018generalization} and has been studied in computer vision~\citep{uselis2025does} and large language models~\citep{sakai2025revisiting}.
In robot learning, prior work has studied generalization to new combinations of environmental factors, such as object type and table height~\citep{gao2024efficient}, as well as recombining demonstrated behaviors to execute new tasks~\citep{mandlekar2020learning}.

More closely related to our setting, recent work has shown that VLAs can struggle to generalize to new combinations of previously learned skills~\citep{xieskillnet}.
One line of work composes learned behaviors at deployment, either by interpolating task-specific text representations from different tasks~\citep{li2025vlas} or by learning reusable skills that can be recombined for new task combinations~\citep{zhai2026skillvla}.
Another approach expands the training data by generating execution trajectories for new skill combinations and retaining successful ones as synthetic demonstrations~\citep{peng2026unleashing}.
We instead exploit constituent-skill coverage in existing demonstrations to provide supervision during fine-tuning, without generating execution trajectories for new combinations.

\vspace{-0.15cm}
\subsection{Shortcut Learning and Counterfactual Supervision}
\label{sec:shortcut-counterfactual}

\looseness=-1
\textit{Shortcut learning} occurs when models exploit predictive correlations in the training data that allow accurate predictions with limited reliance on task-relevant information~\citep{geirhos2020shortcut, si2022language, elisha2026concept}.
In robot demonstrations, the task being executed can often be inferred from the visual observation, allowing a fine-tuned policy to fit demonstrated actions with limited reliance on the instruction~\citep{fang2026vision, xu2025bayesvla, lian2026langforce}.
This can produce a \textit{vision shortcut}, in which the policy uses the visual observation as a proxy for the instruction when predicting actions~\citep{xing2025shortcut}.
In our compositional setting, such a shortcut can cause the policy to execute a demonstrated combination associated with similar observations rather than the new combination specified by the instruction.

\vspace{-0.05cm}
\looseness=-1
Counterfactual training has been used to reduce shortcut reliance by varying a task-relevant factor while holding other inputs fixed and supervising models on the resulting examples~\citep{kaushik2020learning, wan2025eliminating}.
The vision shortcut therefore motivates \emph{counterfactual pairs} that keep the observation fixed while changing the instruction.
Such a pair, however, may lack a corresponding demonstrated action target.
Prior work provides action supervision for such changed instructions by synthesizing counterfactual actions using policies trained on simpler instructions~\citep{glossop2025cast} or by constructing action sequences that connect trajectories for the original and changed instructions~\citep{chen2026resteer}.
In our compositional setting, the changed instruction specifies an undemonstrated combination of demonstrated skills, so the skill currently required under that instruction has already been demonstrated.
Rather than constructing an action target for the counterfactual pair, we use demonstrated executions of the required skill to train the policy so that its action prediction under the changed instruction reflects that skill.

\vspace{-0.2cm}
\section{Problem Formulation}
\label{sec:problem}

\vspace{-0.1cm}
We consider a VLA policy $\pi(\mathbf{a}\mid o,\ell)$ that predicts a chunk of $H$ future actions $\mathbf{a}\in\mathbb{R}^{H\times d_a}$ from a visual observation $o$ and a language instruction $\ell$, where $d_a$ denotes the action dimension.
We fine-tune the policy on robot demonstrations of tasks that share a fixed sequence of operations.

\vspace{-0.05cm}
\looseness=-1
Let $(g_1,\ldots,g_K)$ denote this sequence, such as picking followed by placing.
For each operation $g_k$, a task specifies an entity $e_k \in \mathcal{E}_k$, where $\mathcal{E}_k$ denotes the set of possible entities for that operation, such as objects to manipulate or target locations.
We define a skill as an operation--entity pair $(g_k,e_k)$.
Because the operation sequence is fixed, the entity tuple $c=(e_1,\ldots,e_K)$ uniquely determines a task and its sequence of constituent skills.
The language instruction $\ell$ specifies $c$.
When the current operation is $g_k$, we refer to the skill $(g_k,e_k)$ specified by the instruction as the \emph{currently required skill}.
We assume that, for a fixed observation and currently required skill, changing entity assignments for future operations does not change the action required for the current operation.

\vspace{-0.05cm}
Let $\mathcal{D}$ denote the demonstration dataset used for fine-tuning.
For each demonstration $d\in\mathcal{D}$, let $c_d$ denote the entity tuple specified by its instruction.
The demonstration records the execution of the task determined by $c_d$.
It therefore contains an execution of each constituent skill in that task.
We define the set of demonstrated combinations as
\begin{equation}
\label{eq:demonstrated}
\mathcal{C}_{\mathcal{D}} =
\{\,c_d : d\in\mathcal{D}\,\}
\subseteq
\mathcal{E}_1\times\cdots\times\mathcal{E}_K.
\end{equation}
We study generalization to tasks with entity tuples $c\notin\mathcal{C}_{\mathcal{D}}$ for which every constituent skill has been demonstrated in $\mathcal{D}$.
Our goal is to fine-tune the policy on $\mathcal{D}$ to follow instructions for such tasks while maintaining performance on demonstrated combinations.
The challenge is that these skills are demonstrated in other task combinations, potentially under different observations.
Because the same skill may require different actions across observations, these demonstrations do not directly specify the required actions in a new combination.
This motivates reusing supervision from demonstrated skill executions while accounting for how the required action depends on the observation.

\begin{figure}[t]
\captionsetup{font={stretch=1.0}, skip=5pt}
\centering
\includegraphics[width=\linewidth]{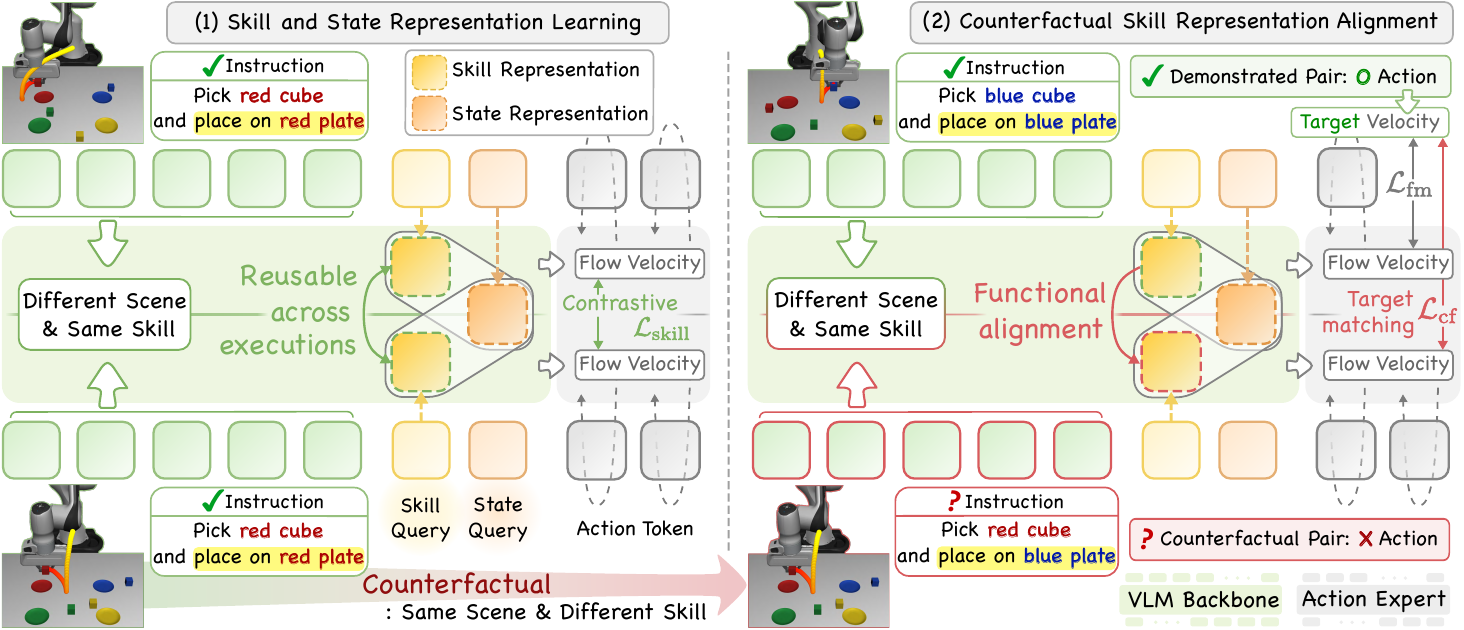}
\caption{\fontsize{9.6}{10.3}\selectfont
Overview of CRAFT.
Yellow instruction highlights mark the currently required skill.
(1) We encourage reuse of skill representations across executions by contrasting prediction errors while holding the state representation and target velocity fixed (target velocity and different-skill replacements omitted for clarity).
(2) We combine the counterfactual skill representation with a state representation from a demonstration of the required skill.
The demonstration's target velocity supervises the prediction, training the counterfactual skill representation to reflect the required skill (skill-changing case shown).
}
\label{fig:craft}
\vspace{-8pt}
\end{figure}

\vspace{-0.1cm}
\section{Method: CRAFT}
\label{sec:method}
\vspace{-0.1cm}
To train the policy on undemonstrated skill combinations, we form counterfactual pairs by holding a demonstration observation fixed while changing the instruction to specify a new combination.
Each pair lacks a corresponding demonstrated action target, even though the currently required skill has been demonstrated.
Actions from those demonstrations cannot generally serve as direct targets
because the same skill can require different actions across observations.
We therefore propose \textbf{CRAFT} (\textbf{C}ounterfactual Skill-\textbf{R}epresentation \textbf{A}lignment for \textbf{F}ine-\textbf{T}uning; Figure~\ref{fig:craft}), which aligns each counterfactual pair's skill representation with the currently required skill under the changed instruction.
The skill representation conditions action prediction on the currently required skill and is trained to be reusable across executions of the same skill.
Instead of using an action from another execution as the counterfactual pair's target, CRAFT uses the pair's skill representation to condition action prediction for a demonstrated execution of the required skill.
We supervise this prediction with that execution's action target, training the pair's skill representation to reflect the required skill.

\vspace{-0.1cm}
\subsection{Skill and State Representations}
\label{sec:method:rep}
\vspace{-0.05cm}
To transfer supervision across executions of the same skill, we introduce a \textit{skill representation} $z^{\text{skill}}$ that reflects the currently required skill and is trained to be reusable across those executions.
Because the same skill may require different actions across observations, we use it with a \textit{state representation} $z^{\text{state}}$ that captures relevant visual information; together, they condition action prediction.

For the VLA architectures we consider~\citep{black2024pi0, black2025pi05, bjorck2025gr00t}, we describe the policy as a pretrained VLM backbone $f_\phi$ coupled with an action expert $p_\theta$.
The backbone produces a vision-language representation $z=f_\phi(o,\ell)$ that conditions the action expert: $\pi(\mathbf{a}\mid o,\ell)=p_\theta(\mathbf{a}\mid z)$.

\looseness=-1
To obtain the skill and state representations, we add two sets of learnable query tokens, $Q^{\text{skill}}$ and $Q^{\text{state}}$, to the VLM token sequence.
Because identifying the currently required skill requires both the instruction and visual context about the current operation, the skill queries attend to both image and text tokens to produce $z^{\text{skill}}$, whereas the state queries attend only to image tokens to produce $z^{\text{state}}$ (Appendix~\ref{app:arch}).
The action expert is then conditioned on both representations:
\begin{equation}
\label{eq:laq}
\pi(\mathbf{a}\mid o,\ell) =
p_\theta(\mathbf{a}\mid z^{\text{skill}},z^{\text{state}}),
\qquad
(z^{\text{skill}},z^{\text{state}}) =
f_\phi(o,\ell;Q^{\text{skill}},Q^{\text{state}}).
\end{equation}

\subsection{Learning Skill Representations from Demonstrations}
\label{sec:method:skill}
\vspace{-0.05cm}
To train $z^{\text{skill}}$ to reflect the currently required skill, we encourage two complementary properties.
A skill representation should be reusable for action prediction across executions of the same skill.
At the same time, representations of different skills should remain distinguishable in action prediction.
Together, these properties motivate a contrastive objective that compares action-prediction errors obtained using skill representations from same- and different-skill executions.

\looseness=-1
Each training sample $i$ consists of an observation $o_i$, an instruction $\ell_i$, and a demonstrated action chunk $\mathbf{a}_i$.
For the VLA models we consider, the action expert $p_\theta$ is trained with flow matching~\citep{lipman2022flow}; we denote its velocity prediction function by $v_\theta$.
Given a noisy action chunk $x_i^t$ at flow time $t$, $v_\theta$ is trained to predict the target velocity $u_i$.
For fixed $i$ and flow time $t$, we replace only its skill representation with $z_j^{\text{skill}}$ from another execution.
Thus, only the skill representation varies across the predictions being compared.
We define the resulting velocity prediction $v_{i\leftarrow j}$ and score $s_{i\leftarrow j}$ as
\begin{equation}
\label{eq:skill-score}
v_{i\leftarrow j} =
v_\theta
\left(
x_i^t,t;
z_i^{\text{state}},
z_j^{\text{skill}}
\right),
\qquad
s_{i\leftarrow j} = -\frac{
\left\|v_{i\leftarrow j}-u_i\right\|_2^2
}{
\sigma_i
},
\end{equation}
where $\sigma_i$ normalizes the prediction-error scale for sample $i$ (Appendix~\ref{app:train:rep}).
A larger $s_{i\leftarrow j}$ therefore indicates lower error in predicting the target velocity $u_i$.

For a sample $i$, the score should be higher for a skill representation from another execution of the same skill than for one from an execution of a different skill.
This comparison requires a label indicating the skill executed at each training sample.
Because the demonstrations do not provide these labels, we use a VLM to segment each demonstration according to the skill being executed and assign the corresponding label to each sample (Appendix~\ref{app:cf}).
For each $i$, we sample a positive $p$ from another execution of the same skill and a set of negatives $\mathcal{N}_i$ from executions of different skills.
Using the scores in Equation~\ref{eq:skill-score}, we define the contrastive objective
\begin{equation}
\label{eq:skill}
\mathcal{L}_{\text{skill}}
=
-\mathbb{E}_i
\left[
\log
\frac{\exp(s_{i\leftarrow p}/\tau)}
{\exp(s_{i\leftarrow p}/\tau)
+
\sum_{n\in\mathcal{N}_i}
\exp(s_{i\leftarrow n}/\tau)}
\right],
\end{equation}
where $\tau$ is a temperature.
Thus, $\mathcal{L}_{\mathrm{skill}}$ encourages skill representations to be reusable across same-skill executions and to preserve distinctions between skills in action prediction.

\subsection{Counterfactual Supervision from Demonstrated Skills}
\label{sec:method:align}
For sample $i$, we form a counterfactual pair by keeping its observation $o_i$ fixed and replacing its instruction $\ell_i$ with a counterfactual instruction $\tilde{\ell}_i$ that specifies a skill combination absent from the demonstrations.
We train the counterfactual pair's skill representation to reflect the skill currently required under $\tilde{\ell}_i$ by using it to condition action prediction for a demonstrated execution of that skill.

\looseness=-1
We construct $\tilde{\ell}_i$ by changing the entities specified for the current or future operations.
Assignments for completed operations remain unchanged because their outcomes are already reflected in $o_i$.
We use a reference sample $r$ from the demonstrations to provide the state representation and supervision target.
If only the entities specified for future operations change, the currently required skill remains unchanged.
For this \textit{skill-preserving} counterfactual, we use the original sample as the reference ($r=i$).
If the entity specified for the current operation changes, the currently required skill changes.
For this \textit{skill-changing} counterfactual, we select $r$ from a demonstrated execution of the required skill.

\looseness=-1
For either counterfactual type, we condition the action expert on the counterfactual pair's skill representation and the reference sample's state representation, and supervise the resulting velocity prediction with the reference sample's target velocity $u_r$.
Let $\tilde{i}_{\mathrm{c}}$ and $\tilde{i}_{\mathrm{p}}$ denote the skill-changing and skill-preserving counterfactual pairs of sample $i$, with skill representations $\tilde{z}_{i,\mathrm{c}}^{\text{skill}}$ and $\tilde{z}_{i,\mathrm{p}}^{\text{skill}}$, respectively.
We define the corresponding velocity predictions and counterfactual objective as
\begin{gather}
v_{r\leftarrow\tilde{i}_{\mathrm{c}}}
=
v_\theta
\left(
x_r^t,t;
z_r^{\text{state}},
\tilde{z}_{i,\mathrm{c}}^{\text{skill}}
\right),
\qquad
v_{i\leftarrow\tilde{i}_{\mathrm{p}}}
=
v_\theta
\left(
x_i^t,t;
z_i^{\text{state}},
\tilde{z}_{i,\mathrm{p}}^{\text{skill}}
\right)
\label{eq:cf-pred}
\\[0.2em]
\mathcal{L}_{\mathrm{cf}}
=
\underbrace{
  \mathbb{E}_{i,r}\,
  \bigl\lVert
    v_{r\leftarrow\tilde{i}_{\mathrm{c}}}
    -u_r
  \bigr\rVert_2^2
}_{\mathcal{L}_{\mathrm{change}}}
+
\underbrace{
  \mathbb{E}_{i}\,
  \bigl\lVert
    v_{i\leftarrow\tilde{i}_{\mathrm{p}}}
    -u_i
  \bigr\rVert_2^2
}_{\mathcal{L}_{\mathrm{preserve}}}
\label{eq:cf}.
\end{gather}
Thus, $\mathcal{L}_{\mathrm{cf}}$ trains the counterfactual pair's skill representation to reflect the skill required under $\tilde{\ell}_i$ in action prediction.
In this way, CRAFT extends supervision to combinations absent from the demonstrations without requiring corresponding demonstrated action targets.

\subsection{Training Objective}
\label{sec:method:craft}
\vspace{-0.05cm}
We combine the standard flow-matching loss for demonstrated action chunks with
$\mathcal{L}_{\mathrm{skill}}$ and $\mathcal{L}_{\mathrm{cf}}$.
Using the notation from Section~\ref{sec:method:skill}, the flow-matching loss is
$\mathcal{L}_{\mathrm{fm}}
=
\mathbb{E}_i
\lVert
v_{i\leftarrow i}-u_i
\rVert_2^2$,
where $v_{i\leftarrow i}$ denotes the velocity prediction conditioned on sample $i$'s own skill and state representations and
$u_i$ is the corresponding target velocity.
The total training objective is
\begin{equation}
\label{eq:total}
\mathcal{L}_{\mathrm{total}}
=
\mathcal{L}_{\mathrm{fm}}
+
\beta\,\mathcal{L}_{\mathrm{skill}}
+
\gamma\,\mathcal{L}_{\mathrm{cf}},
\end{equation}
where $\beta$ and $\gamma$ are the weights for
$\mathcal{L}_{\mathrm{skill}}$ and $\mathcal{L}_{\mathrm{cf}}$, respectively.

We use loss-specific gradient routing.
$\mathcal{L}_{\mathrm{fm}}$ updates all trainable components, whereas
$\mathcal{L}_{\mathrm{skill}}$ stops gradients through the state representation and updates the skill-representation pathway and the action expert.
For $\mathcal{L}_{\mathrm{cf}}$, we do not update the action-expert parameters and stop gradients through the reference state representation, while backpropagating through the expert to the counterfactual skill representation.
We optimize all three objectives jointly in a single training stage (Appendix~\ref{app:train:obj}).

\vspace{-0.1cm}
\section{Experiments}
\label{sec:exp}
\vspace{-0.1cm}
\looseness=-1
In simulation, we evaluate whether policies fine-tuned with CRAFT generalize to undemonstrated combinations while maintaining performance on demonstrated ones (Section~\ref{sec:exp:main}).
We then examine whether policy responses to instruction changes during execution reflect the currently required skill and whether skill representations are organized by skill across executions (Section~\ref{sec:exp:analysis}).
We assess how the objectives for learning reusable skill representations and counterfactual supervision contribute to compositional generalization (Section~\ref{sec:exp:ablation}).
Finally, we evaluate CRAFT on a real robot (Section~\ref{sec:exp:real}).

\begin{figure}[t]
\centering
\includegraphics[width=\linewidth]{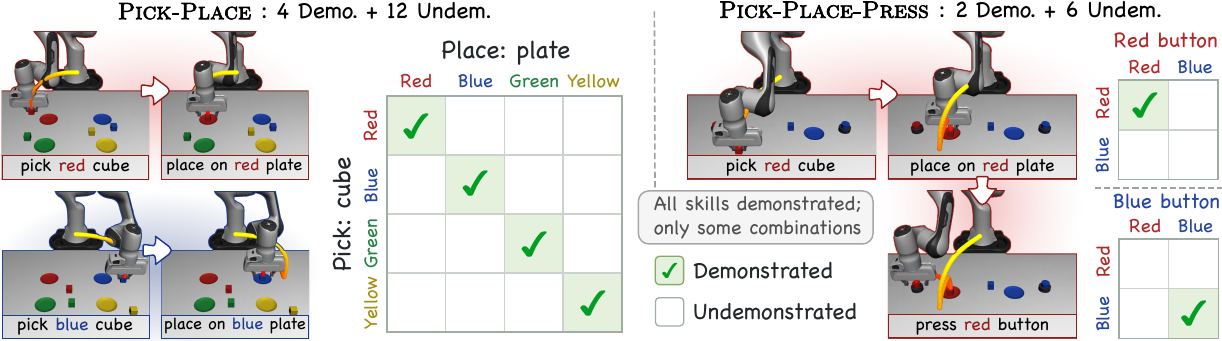}
\captionsetup{font={stretch=1.0}, skip=7pt}
\caption{\fontsize{9.6}{10}\selectfont
Compositional benchmarks with example demonstrations and task-combination matrices (Appendix~\ref{app:benchmark}).
Demonstrations cover all constituent skills but only a subset of combinations (checked cells); empty cells denote undemonstrated combinations.
Rows and columns indicate cube and plate colors, respectively.
The two \textsc{Pick-Place-Press} matrices correspond to red and blue buttons, respectively.
}
\label{fig:benchmark}
\vspace{-10pt}
\end{figure}

\vspace{-0.1cm}
\subsection{Experimental Setup}
\label{sec:exp:setup}

\vspace{-0.1cm}
\paragraph{Benchmarks and task splits.}
We introduce two compositional benchmarks, \textsc{Pick-Place} and \textsc{Pick-Place-Press} (Figure~\ref{fig:benchmark}).
In both benchmarks, the fine-tuning demonstrations cover every constituent skill but only a subset of the possible combinations.
\textsc{Pick-Place} has two operations---picking a cube and placing it on a plate---with four entities per operation, distinguished by color, yielding 16 combinations.
The four same-color combinations are demonstrated; we evaluate the remaining 12.
\textsc{Pick-Place-Press} adds a third operation---pressing a button---and uses two entities per operation, distinguished by color, yielding eight combinations.
The two same-color combinations are demonstrated; we evaluate the remaining six.
We collect 50 demonstrations per demonstrated combination.
We will release both benchmarks and their demonstration datasets.

\vspace{-0.1cm}
\paragraph{Policies and baselines.}
We evaluate CRAFT on three VLA models: $\pi_0$~\citep{black2024pi0}, $\pi_{0.5}$~\citep{black2025pi05}, and GR00T N1.7~\citep{bjorck2025gr00t}.
These models use different VLM--action-expert interfaces: the $\pi$ family uses layer-wise attention to VLM features, whereas GR00T conditions its action expert on the VLM output.
For each model, we compare three standard fine-tuning strategies: \textit{Full} updates both the VLM and action expert, \textit{LoRA}~\citep{hu2021lora} adapts the VLM with LoRA while fully training the action expert, and \textit{Frozen} freezes the VLM and trains only the action expert.
CRAFT uses the \textit{Full} setting.
We additionally compare with CAG-VA~\citep{fang2026vision}, which mitigates vision shortcuts, and a \textit{Skill Composition} baseline inspired by prior work that composes learned behaviors at deployment~\citep{li2025vlas, zhai2026skillvla}.
\textit{Skill Composition} fine-tunes the VLA on constituent skill segments from the demonstrations and sequentially executes the required skills at deployment.
All methods use the same robot demonstrations.

\vspace{-0.15cm}
\paragraph{Evaluation protocol and metrics.}
\looseness=-1
We use task success rate as the primary metric, counting a rollout as successful if the policy completes all stages specified by the instruction within a fixed horizon.
For each combination, 50 initial states are shared across methods and evaluation seeds.
For demonstrated and undemonstrated combinations, we report the mean and standard deviation of success rates over three evaluations of a single trained checkpoint with different environment and action-sampling seeds.
Training, checkpoint selection, and deployment details are
provided in Appendix~\ref{app:implementation}.

\begin{table}[t]
\centering
\captionsetup{font=small, position=top, skip=4pt}
\caption{Success rates (\%) on demonstrated (Demo.) and undemonstrated
(Undem.) combinations across three VLA models and two benchmarks.
We report means and standard deviations over three evaluation seeds.}
\label{tab:main}

\begin{threeparttable}
\small
\setlength{\tabcolsep}{10pt}
\begin{tabular}{lcccccc}
\toprule
&
\multicolumn{2}{c}{$\pi_{0.5}$} &
\multicolumn{2}{c}{$\pi_0$} &
\multicolumn{2}{c}{GR00T N1.7} \\
\cmidrule(lr){2-3}
\cmidrule(lr){4-5}
\cmidrule(lr){6-7}
\bf Method &
\bf Demo. & \bf Undem. &
\bf Demo. & \bf Undem. &
\bf Demo. & \bf Undem. \\
\midrule

\multicolumn{7}{l}{\textsc{Pick--Place}} \\

FT (\textit{Full})
& \ca{$100.0_{\pm0.0}$}
& $8.4_{\pm0.3}$
& \ca{$100.0_{\pm0.0}$}
& $2.8_{\pm0.2}$
& \cc{$98.8_{\pm0.3}$}
& $1.1_{\pm0.1}$ \\

FT (\textit{LoRA})
& \ca{$100.0_{\pm0.0}$}
& \cc{$46.3_{\pm0.9}$}
& \cc{$97.5_{\pm0.0}$}
& $1.0_{\pm0.2}$
& $97.8_{\pm0.6}$
& $2.6_{\pm0.3}$ \\

FT (\textit{Frozen})
& \cb{$99.0_{\pm0.0}$}
& $21.1_{\pm0.2}$
& $48.2_{\pm1.4}$
& $3.0_{\pm0.5}$
& $92.3_{\pm1.3}$
& $2.2_{\pm0.1}$ \\

CAG-VA\tnote{a}
& \cc{$96.3_{\pm1.5}$}
& \cb{$66.5_{\pm0.5}$}
& $50.5_{\pm1.3}$
& \cc{$6.8_{\pm0.6}$}
& $88.7_{\pm1.3}$
& \cc{$4.9_{\pm0.5}$} \\

\textit{Skill Composition}\tnote{b}
& \ca{$100.0_{\pm0.0}$}
& $25.7_{\pm0.4}$
& \cb{$99.5_{\pm0.0}$}
& \cb{$12.7_{\pm0.4}$}
& \ca{$99.7_{\pm0.6}$}
& \cb{$13.1_{\pm0.7}$} \\

\midrule
CRAFT (ours)
& \cb{$99.0_{\pm0.9}$}
& \ca{$84.2_{\pm0.3}$}
& $94.0_{\pm0.9}$
& \ca{$64.3_{\pm0.5}$}
& \cb{$99.0_{\pm0.5}$}
& \ca{$72.1_{\pm0.9}$} \\

\midrule
\midrule

\multicolumn{7}{l}{\textsc{Pick--Place--Press}} \\

FT (\textit{Full})
& \ca{$100.0_{\pm0.0}$}
& $0.7_{\pm0.0}$
& \ca{$99.7_{\pm0.6}$}
& $0.3_{\pm0.0}$
& \cb{$97.7_{\pm2.3}$}
& $1.1_{\pm0.2}$ \\

FT (\textit{LoRA})
& \ca{$100.0_{\pm0.0}$}
& \cc{$15.4_{\pm0.8}$}
& \cb{$99.3_{\pm0.6}$}
& $0.7_{\pm0.0}$
& \ca{$99.3_{\pm1.2}$}
& $1.0_{\pm0.0}$ \\

FT (\textit{Frozen})
& \cc{$98.0_{\pm1.0}$}
& $0.9_{\pm0.2}$
& $82.0_{\pm3.5}$
& \cb{$1.5_{\pm0.4}$}
& \cb{$97.7_{\pm0.6}$}
& $1.1_{\pm0.2}$ \\

CAG-VA\tnote{a}
& $92.7_{\pm3.5}$
& \cb{$21.8_{\pm1.9}$}
& $69.0_{\pm2.6}$
& \cc{$1.1_{\pm0.2}$}
& \cc{$88.7_{\pm2.5}$}
& \cb{$2.1_{\pm0.2}$} \\

\textit{Skill Composition}\tnote{b}
& \cb{$99.7_{\pm0.6}$}
& $5.8_{\pm0.2}$
& \cb{$99.3_{\pm0.6}$}
& $0.3_{\pm0.0}$
& \ca{$99.3_{\pm0.6}$}
& \cc{$2.0_{\pm0.0}$} \\

\midrule
CRAFT (ours)
& \cb{$99.7_{\pm0.6}$}
& \ca{$62.6_{\pm3.9}$}
& \cc{$95.3_{\pm0.6}$}
& \ca{$40.7_{\pm1.5}$}
& \ca{$99.3_{\pm0.6}$}
& \ca{$31.9_{\pm1.1}$} \\

\bottomrule
\end{tabular}

\begin{tablenotes}[flushleft]
\footnotesize
\item[]
For FT, \textit{Full}, \textit{LoRA}, and \textit{Frozen} denote VLM adaptation; the action expert is fully trained in all cases.

\item[a]
For CAG-VA~\citep{fang2026vision}, we tune the guidance strength for
\textit{Full}, \textit{LoRA}, and \textit{Frozen} and report the selected
configuration for each VLA model and benchmark; selection details are
provided in Appendix~\ref{app:results:cagva}.

\item[b]
\textit{Skill Composition} uses the \textit{Full} setting to fine-tune each VLA on constituent skill segments and sequentially executes the required skills at deployment.
\end{tablenotes}
\vspace{-0.4cm}
\end{threeparttable}
\end{table}

\vspace{-0.15cm}
\subsection{Compositional Generalization}
\label{sec:exp:main}
Table~\ref{tab:main} reports success rates on demonstrated and undemonstrated combinations.
Across all VLA models and benchmarks, CRAFT achieves the highest success on undemonstrated combinations while maintaining high performance on demonstrated ones.
By contrast, standard fine-tuning often performs well on demonstrated combinations but achieves substantially lower success on the remaining combinations.
This gap indicates that high success on demonstrated combinations does not by itself imply the ability to recombine their constituent skills.

\looseness=-1
Changing the adaptation method alone does not close this gap.
The gap persists across \textit{Full}, \textit{LoRA}, and \textit{Frozen}; \textit{Full} has low success on these combinations despite updating the entire VLM.
Consistent with our vision-shortcut hypothesis, this may arise because policies can fit demonstrated actions using observation--action correlations that do not generalize to new skill combinations.

CAG-VA and \textit{Skill Composition} represent two alternatives: strengthening language conditioning and composing learned skills at deployment.
CAG-VA improves generalization in several settings, but its gains vary substantially across models and benchmarks.
\textit{Skill Composition} improves over \textit{Full} on \textsc{Pick-Place} for all three models, but its gains are much smaller on \textsc{Pick-Place-Press}.

In contrast, CRAFT's improvement over the baselines persists across all three VLA models on \textsc{Pick-Place-Press}, suggesting that its gains are not specific to \textsc{Pick-Place}.
Nevertheless, a substantial gap remains between demonstrated and undemonstrated performance, particularly on \textsc{Pick-Place-Press}.
We provide detailed compositional-generalization results in Appendix~\ref{app:results:combinations}.

\begin{figure}[t]
\centering
\includegraphics[width=\linewidth]{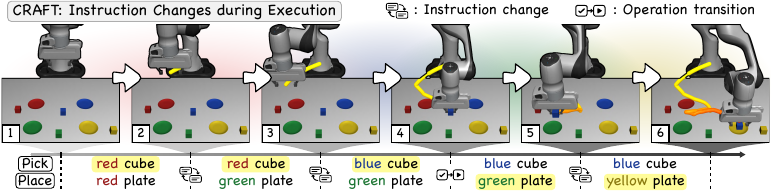}
\captionsetup{font={stretch=1.0}, skip=7pt}
\caption{\fontsize{9.8}{10}\selectfont
Responses to instruction changes.
Yellow highlights identify the entity assigned to the current operation.
The robot approaches the red cube and maintains this behavior when only the future placement target changes (1$\rightarrow$2$\rightarrow$3).
It redirects its approach when the picking target changes (3$\rightarrow$4), then changes its destination during placement (5$\rightarrow$6), placing the blue cube on the yellow plate.
}
\label{fig:behavior}
\vspace{-10pt}
\end{figure}

\vspace{-0.2cm}
\subsection{Behavior and Representation Analysis}
\label{sec:exp:analysis}
\begin{wrapfigure}{r}{0.52\linewidth}
\vspace{-\intextsep}
\centering
\includegraphics[width=\linewidth]{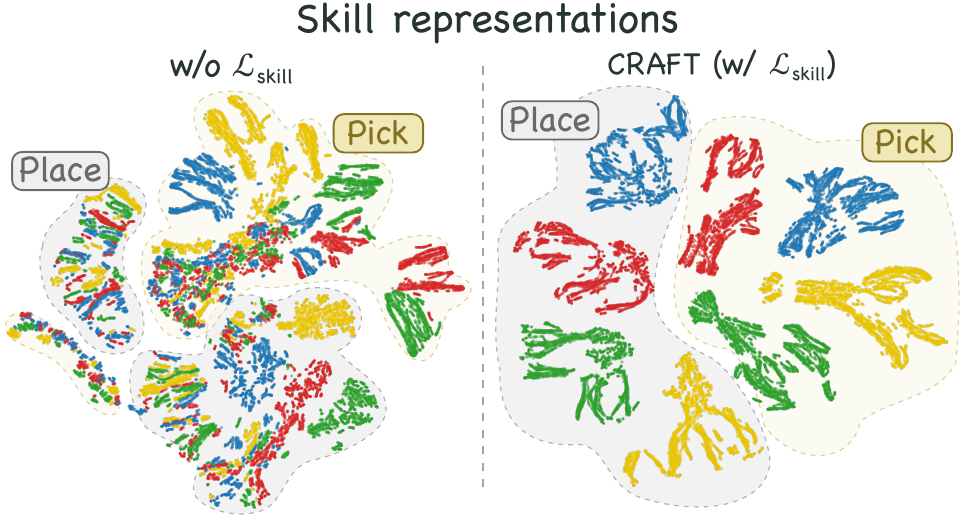}
\captionsetup{font={stretch=1.0}, skip=6pt}  
\caption{\fontsize{9.3}{10}\selectfont
Skill representations learned with and without $\mathcal{L}_{\mathrm{skill}}$ on \textsc{Pick-Place}, independently projected with t-SNE.
Each point represents a demonstration frame, colored by the entity of the currently required skill.
Shaded regions indicate pick and place groups.
}
\label{fig:representation}
\vspace{-\intextsep}
\end{wrapfigure}
\looseness=-1
We examine whether policy responses to instruction changes during execution reflect the currently required skill (Figure~\ref{fig:behavior}; Appendix~\ref{app:results:instruction-response}).
When an instruction change leaves the currently required skill unchanged, the policy continues executing that skill.
When the required skill changes, the policy begins executing the newly required skill.
Together, these responses suggest that the policy remains responsive to the instruction throughout execution.
Notably, it does so from a state reached while executing a different skill.
This suggests the policy can execute the required skill from states reached during other skill executions, rather than only from states encountered in demonstrations of that skill.

\vspace{-0.05cm}
\looseness=-1
Figure~\ref{fig:representation} examines how $\mathcal{L}_{\mathrm{skill}}$ shapes the learned skill representations.
Without $\mathcal{L}_{\mathrm{skill}}$, representations from different entities are more mixed within the pick and place groups.
With $\mathcal{L}_{\mathrm{skill}}$, they become more clearly organized by the currently required skill, forming entity-specific groups within each operation.
This organization is consistent with $\mathcal{L}_{\mathrm{skill}}$ encouraging skill-dependent structure in the learned representations.
We evaluate whether these representations are functionally reusable across executions in Appendix~\ref{app:results:skill-swap}.

\vspace{-0.23cm}
\subsection{Training Objective Ablations}
\label{sec:exp:ablation}
\vspace{-0.1cm}
We compare combinations of auxiliary objectives while keeping the architecture and flow-matching loss unchanged (Table~\ref{tab:ablation}).
Success on undemonstrated combinations is low for both models with $\mathcal{L}_{\mathrm{fm}}$ alone and remains low when $\mathcal{L}_{\mathrm{skill}}$ is added.
This objective encourages skill representations to be reusable across same-skill executions, but does not directly supervise representations induced by counterfactual instructions.
Under this setting, adding $\mathcal{L}_{\mathrm{preserve}}$ provides no clear improvement, whereas $\mathcal{L}_{\mathrm{change}}$ improves success on undemonstrated combinations for both models.

\begin{wraptable}{r}{0.54\linewidth}
\vspace{-\intextsep}
\captionsetup{font=small, position=top, skip=4pt}
\caption{Success rates (\%) on \textsc{Pick-Place} for training objective ablations. (mean $\pm$ std.; three evaluation seeds)}
\label{tab:ablation}
\centering

\fontsize{7.5}{9}\selectfont
\setlength{\tabcolsep}{2.5pt}

\begin{tabular*}{\linewidth}{@{\extracolsep{\fill}}ccc cccc}
\toprule
\multicolumn{3}{c}{Auxiliary objectives}
& \multicolumn{2}{c}{$\pi_{0.5}$}
& \multicolumn{2}{c}{GR00T N1.7} \\
\cmidrule(lr){1-3}
\cmidrule(lr){4-5}
\cmidrule(lr){6-7}

\bf Skill & \bf Pres. & \bf Change
& \bf Demo. & \bf Undem.
& \bf Demo. & \bf Undem. \\
\midrule

&
&
&
$88.8_{\pm0.6}$
& \cc{$1.9_{\pm0.4}$}
& \ca{$99.3_{\pm0.3}$}
& $3.8_{\pm0.1}$ \\

$\checkmark$
&
&
&
\cb{$99.2_{\pm0.6}$}
& $0.3_{\pm0.4}$
& $97.2_{\pm1.0}$
& \cc{$6.1_{\pm0.5}$} \\

$\checkmark$
& $\checkmark$
&
&
\ca{$99.8_{\pm0.3}$}
& $0.4_{\pm0.2}$
& \cc{$97.3_{\pm0.6}$}
& $5.3_{\pm0.3}$ \\

$\checkmark$
&
& $\checkmark$
&
$97.7_{\pm0.3}$
& \cb{$14.7_{\pm0.8}$}
& $94.5_{\pm1.3}$
& \cb{$12.1_{\pm0.4}$} \\

\midrule

\multicolumn{3}{l}{\textbf{CRAFT (all)}}
& \cc{$99.0_{\pm0.9}$}
& \ca{$84.2_{\pm0.3}$}
& \cb{$99.0_{\pm0.5}$}
& \ca{$72.1_{\pm0.9}$} \\

\bottomrule
\end{tabular*}

\vspace{2pt}
\parbox{\linewidth}{\scriptsize
All variants include $\mathcal{L}_{\mathrm{fm}}$.
\textbf{Skill}, \textbf{Pres.}, and \textbf{Change} denote
$\mathcal{L}_{\mathrm{skill}}$,
$\mathcal{L}_{\mathrm{preserve}}$, and
$\mathcal{L}_{\mathrm{change}}$, respectively.
CRAFT uses all three auxiliary objectives.}

\vspace{-\intextsep}
\end{wraptable}

\vspace{-0.05cm}
Combining $\mathcal{L}_{\mathrm{change}}$ and $\mathcal{L}_{\mathrm{preserve}}$ yields substantially higher success on undemonstrated combinations than using either counterfactual term separately.
The combined objective also maintains high success on demonstrated combinations.
These results suggest that counterfactual supervision benefits from covering both instruction changes that alter the currently required skill and those that leave it unchanged.
We provide additional ablations and sensitivity analyses in Appendix~\ref{app:results:queries}.

\begin{figure}[t]
\centering
\captionsetup{font={stretch=1.0}, skip=4pt}
\includegraphics[width=\linewidth]{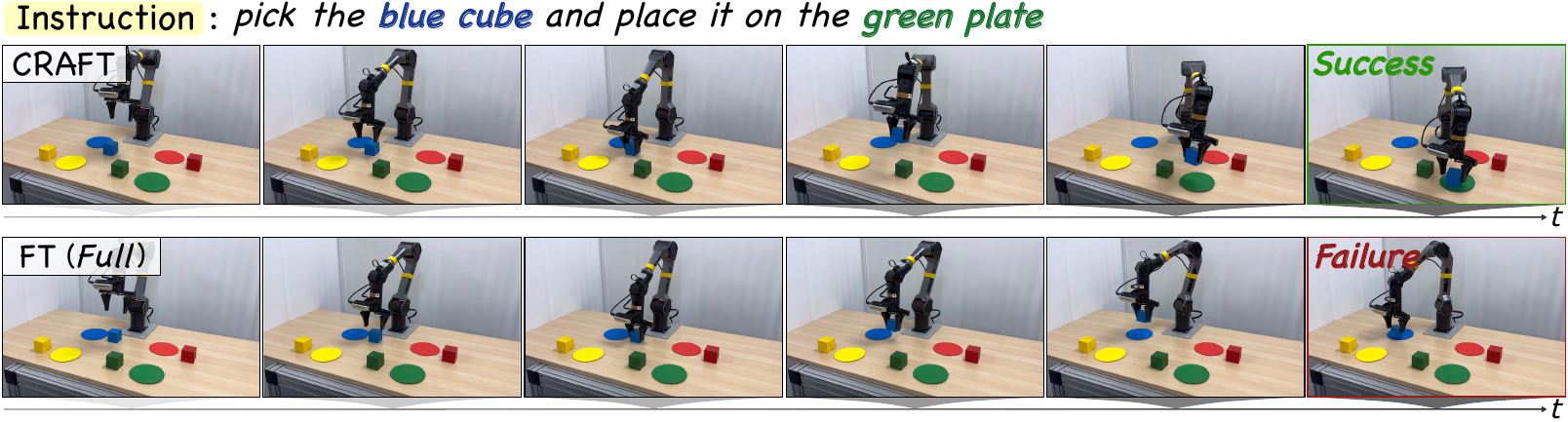}
\caption{\fontsize{10}{10}\selectfont
Real-robot execution of an undemonstrated combination.
The instruction requires picking the blue cube and placing it on the green plate; this combination is absent from the demonstrations, although both constituent skills have been demonstrated.
Both CRAFT and \textit{Full} pick the instructed blue cube, but CRAFT places it on the instructed green plate, whereas \textit{Full} places it on the blue plate, corresponding to the demonstrated blue-to-blue combination.
}
\vspace{-10pt}
\label{fig:real}
\end{figure}

\subsection{Real-Robot Evaluation}
\label{sec:exp:real}
\vspace{-0.1cm}
We evaluate whether CRAFT improves compositional generalization on a real robot.
The real-robot setup follows the task structure of \textsc{Pick-Place}, with four differently colored cubes and four plates of the same four colors.
A Piper 6-DoF robot performs the tasks using observations from third-person and wrist cameras.
The demonstration dataset contains 20 demonstrations for each of the four same-color combinations, leaving the remaining 12 combinations undemonstrated.
We compare CRAFT with FT (\textit{Full}) on $\pi_{0.5}$; both use the same demonstration dataset and are evaluated over five trials per combination.

\begin{wraptable}{r}{0.43\linewidth}
\vspace{-\intextsep}
\captionsetup{size=small, position=top, skip=7pt}
\caption{Real-robot task success (successful trials / total trials) on demonstrated (Demo.) and undemonstrated (Undem.) combinations.}
\label{tab:real}
\centering
\small
\begin{tabular*}{\linewidth}{@{\extracolsep{\fill}}lcc}
\toprule
\textbf{Method} & \textbf{Demo.} & \textbf{Undem.} \\
\midrule
FT (\textit{Full})  & \ca{19/20} & \cc{9/60} \\
CRAFT (ours)  & \ca{19/20} & \ca{43/60} \\
\bottomrule
\end{tabular*}
\vspace{-\intextsep}
\end{wraptable}

Figure~\ref{fig:real} provides a qualitative comparison of \textit{Full} and CRAFT on an undemonstrated combination.
CRAFT achieves higher success on undemonstrated combinations than \textit{Full} while maintaining high success on demonstrated combinations (Table~\ref{tab:real}).
These results suggest that CRAFT can improve real-robot compositional generalization when demonstrations cover the constituent skills but not every combination.
We provide additional real-robot rollouts in Appendix~\ref{app:results:real-rollouts}.

\vspace{-0.15cm}
\section{Discussion}
\label{sec:discussion}

\vspace{-0.18cm}
\paragraph{Where the supervision comes from.}
Our setting lacks demonstrations for some skill combinations, but each constituent skill has already been demonstrated.
CRAFT therefore uses demonstrated executions of the currently required skill as the source of supervision, rather than synthesizing trajectories for new combinations or transferring actions directly across observations.

\vspace{-0.18cm}
\paragraph{What the results suggest.}
High success on demonstrated combinations did not consistently translate into high success on undemonstrated combinations.
The gains from CRAFT suggest that existing demonstrations can provide supervision beyond the skill combinations they contain.

\vspace{-0.18cm}
\paragraph{Limitations.}
Our formulation assumes a fixed operation sequence and that every constituent skill in an undemonstrated combination appears in the demonstrations.
Tasks requiring new skills or operation sequences that vary with the instruction are outside the scope of our formulation.

\vspace{-0.15cm}
\section{Conclusion}
\label{sec:conclusion}
\vspace{-0.1cm}
We studied whether VLAs fine-tuned on demonstrations can recombine demonstrated skills into combinations absent from those demonstrations.
Motivated by the vision shortcut that can arise during fine-tuning, we proposed CRAFT, which uses counterfactual supervision to train the policy to follow the currently required skill without requiring demonstrations of every skill combination.
Across three VLA models and two simulation benchmarks, CRAFT improved success on undemonstrated combinations while maintaining high performance on demonstrated ones.
CRAFT also improved compositional generalization on a real robot.
These results show that compositional generalization can be improved by making better use of constituent-skill supervision present in demonstrations.

\section*{Ethics Statement}

This work studies compositional generalization for vision-language-action policies in simulation and on a real robot.
Our experiments do not involve human or animal subjects, personally identifiable information, or sensitive personal data.

We use publicly released VLA models~\citep{black2024pi0, black2025pi05, pertsch2025fast, bjorck2025gr00t} and construct our simulation benchmarks by adapting the LIBERO environment~\citep{liu2023libero} to the compositional setting studied in this work.
The demonstration datasets for the simulation benchmarks and real-robot experiments consist of robot manipulation trajectories collected for the corresponding tasks.

The real-robot experiments are conducted in a controlled laboratory setting and evaluate compositional generalization rather than safety for unconstrained real-world deployment.

\section*{Reproducibility Statement}

To support reproducibility, we provide implementation details in the main paper and additional architecture details, training configurations, and hyperparameters in the appendix.
We also specify the evaluation protocols used for the reported experiments.
The datasets and evaluation environments developed for this work, together with the corresponding training code, will be publicly released.

\section*{AI Use Disclosure}

The authors used generative AI tools for feedback on research methodology and experiments, assistance with code debugging, and manuscript editing for clarity.
All AI-assisted content was reviewed and finalized by the authors.
The authors take full responsibility for the final content of this work.

\bibliography{iclr2027_conference}
\bibliographystyle{iclr2027_conference}

\appendix
\appendix
\counterwithin{figure}{section}
\counterwithin{table}{section}

\section{Motivating Experiments}
\label{app:motivation}

\subsection{Compositional Generalization Gap in Fine-Tuned VLA Models}
\label{app:compositional-failure}

\begin{figure}[!ht]
\begin{center}
  \includegraphics[width=0.9\linewidth]{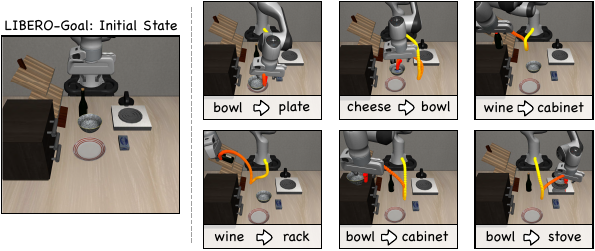}
\end{center}
\caption{
An example initial scene and the six pick-and-place tasks from the LIBERO-\textsc{Goal} suite.
The first panel shows the initial scene, and the remaining panels show the six tasks with their instructions.
}
\label{fig:libero-scene}
\end{figure}

\paragraph{Setup.}
LIBERO~\citep{liu2023libero} is a simulation benchmark for language-conditioned manipulation and has been used to evaluate recent VLA models~\citep{kim2024openvla, black2024pi0, black2025pi05, bjorck2025gr00t}.
The LIBERO-\textsc{Goal} suite contains ten tasks that share the same scene and initial-state distribution but differ in their goals.
We focus on its six pick-and-place tasks for compositional evaluation; the remaining four tasks involve other manipulation objectives.
Figure~\ref{fig:libero-scene} shows an example initial scene and the six pick-and-place tasks.

\paragraph{Constructing undemonstrated combinations.}
We represent each pick-and-place task by its object--target pair, corresponding to a pick skill and a place skill.
The six original tasks define the demonstrated combinations marked with $\checkmark$ in Table~\ref{tab:libero-grid}.
The remaining nine object--target pairs are undemonstrated combinations, while every constituent pick and place skill appears in at least one demonstrated task.
We manually select five of these nine combinations before model evaluation, excluding cases in which task feasibility or manipulation difficulty could confound compositional generalization.
To avoid introducing new instruction phrasing, each constructed instruction reuses object and target phrases from the demonstrated pick-and-place tasks (Table~\ref{tab:libero-ood}).
For each constructed task, we define a new Behavior Domain Definition Language (BDDL) specification with the corresponding goal condition while preserving the suite's scene and initial-state distribution.

\begin{table}[!ht]
\caption{
Object--target combinations derived from the six pick-and-place tasks in LIBERO-\textsc{Goal}.
$\checkmark$ marks a demonstrated combination, with the corresponding task index from Table~\ref{tab:libero-id};
$\circ$ marks an undemonstrated combination used for evaluation.
Blank cells denote undemonstrated combinations not evaluated.
}
\label{tab:libero-grid}
\begin{center}
\small
\begin{tabular}{lccccc}
\toprule
& \bf cabinet top & \bf rack & \bf plate & \bf stove & \bf bowl \\
\midrule
wine bottle   & $\checkmark$\,(1) & $\checkmark$\,(4) & $\circ$ & $\circ$ & \\
bowl          & $\checkmark$\,(2) & & $\checkmark$\,(3) & $\checkmark$\,(6) & \\
cream cheese  & $\circ$ & & $\circ$ & $\circ$ & $\checkmark$\,(5) \\
\bottomrule
\end{tabular}
\end{center}
\end{table}

\paragraph{Models and evaluation.}
We evaluate publicly released LIBERO-fine-tuned checkpoints of $\pi_0$~\citep{black2024pi0}, $\pi_{0.5}$~\citep{black2025pi05}, $\pi_0$-FAST~\citep{pertsch2025fast}, and GR00T N1.7~\citep{bjorck2025gr00t}.
We use each checkpoint without further training.
For each model and task, we evaluate $50$ trials for each of three evaluation seeds, sampling initial states according to the LIBERO-\textsc{Goal} protocol.
Each policy executes the first five actions of each predicted chunk before replanning.

\paragraph{Results.}
Across the four models, mean success on the six demonstrated combinations ranges from $88.8$ to $97.4$\%, compared with $2.0$ to $50.9$\% on the five undemonstrated combinations (Tables~\ref{tab:libero-id} and~\ref{tab:libero-ood}).
The resulting demonstrated-to-undemonstrated gap ranges from $46.5$ to $94.1$ percentage points.
Although the magnitude of the gap varies across models, all four show substantially lower success on undemonstrated combinations despite every constituent skill appearing in the demonstrated tasks.
We analyze the associated failure behavior further in Appendix~\ref{app:vision-shortcut}.

\begin{table}[t]
\caption{
Success rate (\%) on the ten original LIBERO-\textsc{Goal} tasks.
Released checkpoints are evaluated on $50$ trials per task and evaluation seed; we report mean $\pm$ std over three seeds.
Task indices are used for cross-reference with Tables~\ref{tab:libero-grid} and~\ref{tab:libero-ood}.
}
\label{tab:libero-id}
\begin{center}
\small
\begin{tabular}{clcccc}
\toprule
& \bf Task & $\pi_0$ & $\pi_{0.5}$ & $\pi_0$-FAST & GR00T N1.7 \\
\midrule
\multicolumn{6}{l}{\textit{pick-and-place}} \\
1  & put the wine bottle on top of the cabinet   & $98.0_{\pm1.6}$  & $98.0_{\pm1.6}$  & $97.3_{\pm0.9}$  & $98.7_{\pm0.9}$ \\
2  & put the bowl on top of the cabinet          & $97.3_{\pm1.9}$  & $96.7_{\pm0.9}$  & $93.3_{\pm6.6}$  & $94.7_{\pm1.9}$ \\
3  & put the bowl on the plate                   & $100.0_{\pm0.0}$ & $99.3_{\pm0.9}$  & $98.0_{\pm0.0}$  & $96.0_{\pm1.6}$ \\
4  & put the wine bottle on the rack             & $75.3_{\pm1.9}$  & $95.3_{\pm2.5}$  & $46.0_{\pm1.6}$  & $96.0_{\pm1.6}$ \\
5  & put the cream cheese in the bowl            & $98.7_{\pm0.9}$  & $98.0_{\pm0.0}$  & $99.3_{\pm0.9}$  & $93.3_{\pm0.9}$ \\
6  & put the bowl on the stove                   & $98.0_{\pm1.6}$  & $97.3_{\pm0.9}$  & $98.7_{\pm0.9}$  & $98.0_{\pm1.6}$ \\
\cmidrule(lr){1-6}
   & \textit{mean}                               & $94.6_{\pm0.8}$  & $97.4_{\pm0.6}$  & $88.8_{\pm1.0}$  & $96.1_{\pm0.2}$ \\
\midrule
\multicolumn{6}{l}{\textit{other}} \\
7  & open the top drawer and put the bowl inside & $82.0_{\pm4.3}$  & $93.3_{\pm2.5}$  & $88.7_{\pm3.4}$  & $86.7_{\pm2.5}$ \\
8  & turn on the stove                           & $99.3_{\pm0.9}$  & $100.0_{\pm0.0}$ & $94.0_{\pm0.0}$  & $100.0_{\pm0.0}$ \\
9  & open the middle drawer of the cabinet       & $98.7_{\pm0.9}$  & $96.7_{\pm0.9}$  & $74.7_{\pm2.5}$  & $95.3_{\pm3.4}$ \\
10 & push the plate to the front of the stove    & $93.3_{\pm2.5}$  & $94.0_{\pm2.8}$  & $87.3_{\pm3.4}$  & $92.0_{\pm1.6}$ \\
\cmidrule(lr){1-6}
   & \textit{mean}                               & $93.3_{\pm1.4}$  & $96.0_{\pm1.1}$  & $86.2_{\pm1.2}$  & $93.5_{\pm1.1}$ \\
\bottomrule
\end{tabular}
\end{center}
\end{table}

\begin{table}[t]
\caption{
Success rate (\%) on the five constructed undemonstrated combinations.
We report mean $\pm$ std over three evaluation seeds, with $50$ trials per task and seed.
}
\label{tab:libero-ood}
\begin{center}
\small
\begin{tabular}{lcccc}
\toprule
\bf Instruction & $\pi_0$ & $\pi_{0.5}$ & $\pi_0$-FAST & \bf GR00T N1.7 \\
\midrule
put the wine bottle on the plate           & $6.0_{\pm0.0}$ & $52.0_{\pm3.3}$ & $0.0_{\pm0.0}$  & $0.0_{\pm0.0}$ \\
put the wine bottle on the stove           & $2.0_{\pm1.6}$ & $77.3_{\pm4.1}$ & $0.7_{\pm0.9}$  & $0.0_{\pm0.0}$ \\
put the cream cheese on top of the cabinet & $0.0_{\pm0.0}$ & $78.7_{\pm2.5}$ & $7.3_{\pm1.9}$  & $6.7_{\pm2.5}$ \\
put the cream cheese on the plate          & $4.0_{\pm1.6}$ & $24.7_{\pm7.7}$ & $88.0_{\pm1.6}$ & $0.7_{\pm0.9}$ \\
put the cream cheese on the stove          & $2.7_{\pm0.9}$ & $22.0_{\pm2.8}$ & $1.3_{\pm0.9}$  & $2.7_{\pm1.9}$ \\
\midrule
\textit{mean}                              & $2.9_{\pm0.4}$ & $50.9_{\pm2.4}$ & $19.5_{\pm0.7}$ & $2.0_{\pm1.0}$ \\
\bottomrule
\end{tabular}
\end{center}
\end{table}

\subsection{Failure Behavior Consistent with a Vision Shortcut}
\label{app:vision-shortcut}

Appendix~\ref{app:compositional-failure} shows a substantial gap between demonstrated and undemonstrated combinations, even when every constituent skill has been demonstrated.
We next examine how the policies fail on undemonstrated combinations and whether their behavior is consistent with the vision-shortcut hypothesis.

\begin{figure}[!ht]
\centering
\includegraphics[width=0.8\linewidth]{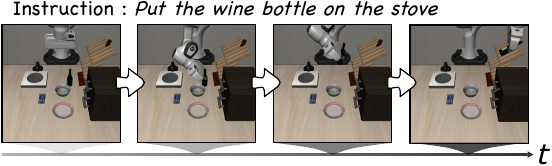}
\caption{
Failure on an undemonstrated combination in LIBERO-\textsc{Goal}.
Given ``put the wine bottle on the stove,'' the fine-tuned $\pi_{0.5}$ policy grasps the instructed object but places it on the rack, corresponding to the demonstrated ``put the wine bottle on the rack'' task.
}
\label{fig:libero-vision-shortcut}
\end{figure}

\paragraph{A representative failure on an undemonstrated combination.}
Figure~\ref{fig:libero-vision-shortcut} shows a rollout from the fine-tuned $\pi_{0.5}$ policy under the undemonstrated instruction ``put the wine bottle on the stove.''
Although this object--target combination is absent from the demonstrations, both constituent skills---picking the wine bottle and placing an object on the stove---have been demonstrated.
The policy grasps the instructed wine bottle but places it on the rack rather than the stove.
This outcome corresponds to the demonstrated task ``put the wine bottle on the rack,'' illustrating a failure in which the policy completes a familiar demonstrated combination instead of the instructed one.

\paragraph{Which demonstrated tasks failed rollouts resemble.}
To determine whether this behavior extends beyond a single example, we analyze the outcomes of rollouts on the five constructed undemonstrated combinations from Appendix~\ref{app:compositional-failure}.
For each rollout, we evaluate the goal predicates of the ten original LIBERO-\textsc{Goal} tasks and record whether it completes (i) the instructed undemonstrated task, (ii) a demonstrated task that shares the object with the instruction, (iii) a demonstrated task that shares the target, (iv) another original task, or (v) none of them.
If multiple task predicates become satisfied during a rollout, we assign the rollout to the task whose predicate is satisfied first.
Table~\ref{tab:libero-outcome} summarizes the results.

Across models, when a rollout completes one of the original demonstrated tasks instead of the instructed combination, it almost always completes a task that shares either the object or the target with the instruction.
The dominant pattern differs across models: $\pi_{0.5}$ and $\pi_0$-FAST more often complete object-sharing tasks, whereas $\pi_0$ more often completes target-sharing tasks; GR00T N1.7 exhibits both patterns.
Rollouts categorized as \textit{None} satisfy none of the evaluated task goals.
In several manually inspected None rollouts, the failure occurred earlier in execution, such as when the policy failed to grasp the instructed object.
Thus, when failed rollouts do complete an evaluated task, their outcomes are strongly concentrated on demonstrated combinations that partially match the instruction.

\begin{table}[!ht]
\caption{
Outcomes of rollouts on the five constructed undemonstrated combinations.
Values are percentages of trials pooled across combinations, reported as mean $\pm$ std over three seeds.
\textit{None} indicates that the rollout satisfies neither the instructed task goal nor any of the ten original LIBERO-\textsc{Goal} task predicates.
}
\label{tab:libero-outcome}
\begin{center}
\small
\begin{tabular}{lccccc}
\toprule
\bf Model & \bf Instructed task & \bf Shares the object & \bf Shares the target & \bf Other of the ten & \bf None \\
\midrule
$\pi_0$      & $2.9_{\pm0.4}$  & $9.6_{\pm0.7}$  & $48.8_{\pm0.6}$ & $0.5_{\pm0.4}$ & $38.1_{\pm1.2}$ \\
$\pi_{0.5}$  & $50.9_{\pm2.4}$ & $38.4_{\pm1.2}$ & $2.8_{\pm0.3}$  & $0.0_{\pm0.0}$ & $7.9_{\pm0.9}$  \\
$\pi_0$-FAST & $19.5_{\pm0.7}$ & $41.9_{\pm2.0}$ & $4.1_{\pm0.2}$  & $0.8_{\pm0.7}$ & $33.7_{\pm0.8}$ \\
GR00T N1.7   & $2.0_{\pm1.0}$  & $39.3_{\pm1.0}$ & $25.5_{\pm1.6}$ & $1.9_{\pm0.7}$ & $31.3_{\pm1.0}$ \\
\bottomrule
\end{tabular}
\end{center}
\end{table}

\begin{figure}[!ht]
\centering
\includegraphics[width=0.8\linewidth]{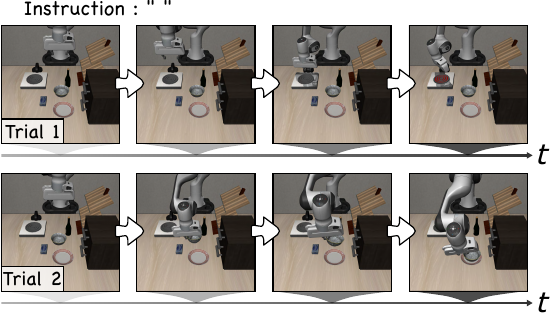}
\caption{
Behavior of the fine-tuned $\pi_{0.5}$ policy under an empty instruction.
Without a task instruction, the policy turns on the stove in one trial (top) and puts the bowl on the plate in another (bottom), both corresponding to demonstrated tasks.
}
\label{fig:libero-empty-instruction}
\end{figure}

\paragraph{Behavior under an empty instruction.}
As a complementary qualitative probe, we evaluate the same fine-tuned $\pi_{0.5}$ policy with an empty instruction.
Figure~\ref{fig:libero-empty-instruction} shows two example trials.
Despite receiving no task instruction, the policy executes behaviors corresponding to demonstrated tasks: it turns on the stove in one trial and puts the bowl on the plate in the other.
Although an empty instruction is outside the standard task setting, these examples show that demonstrated-task behaviors can arise even without an explicit task instruction.

\paragraph{Summary.}
On undemonstrated combinations, failed rollouts that complete an evaluated task are strongly biased toward demonstrated combinations sharing a constituent with the instruction, while demonstrated-task behaviors can also arise under an empty instruction.
These observations provide behavioral evidence for the vision-shortcut failure mode, while not excluding other contributing factors.

\section{Method Details}
\label{app:method}

\subsection{Architecture}
\label{app:arch}

\begin{figure}[t]
\centering
\includegraphics[width=0.85\linewidth]{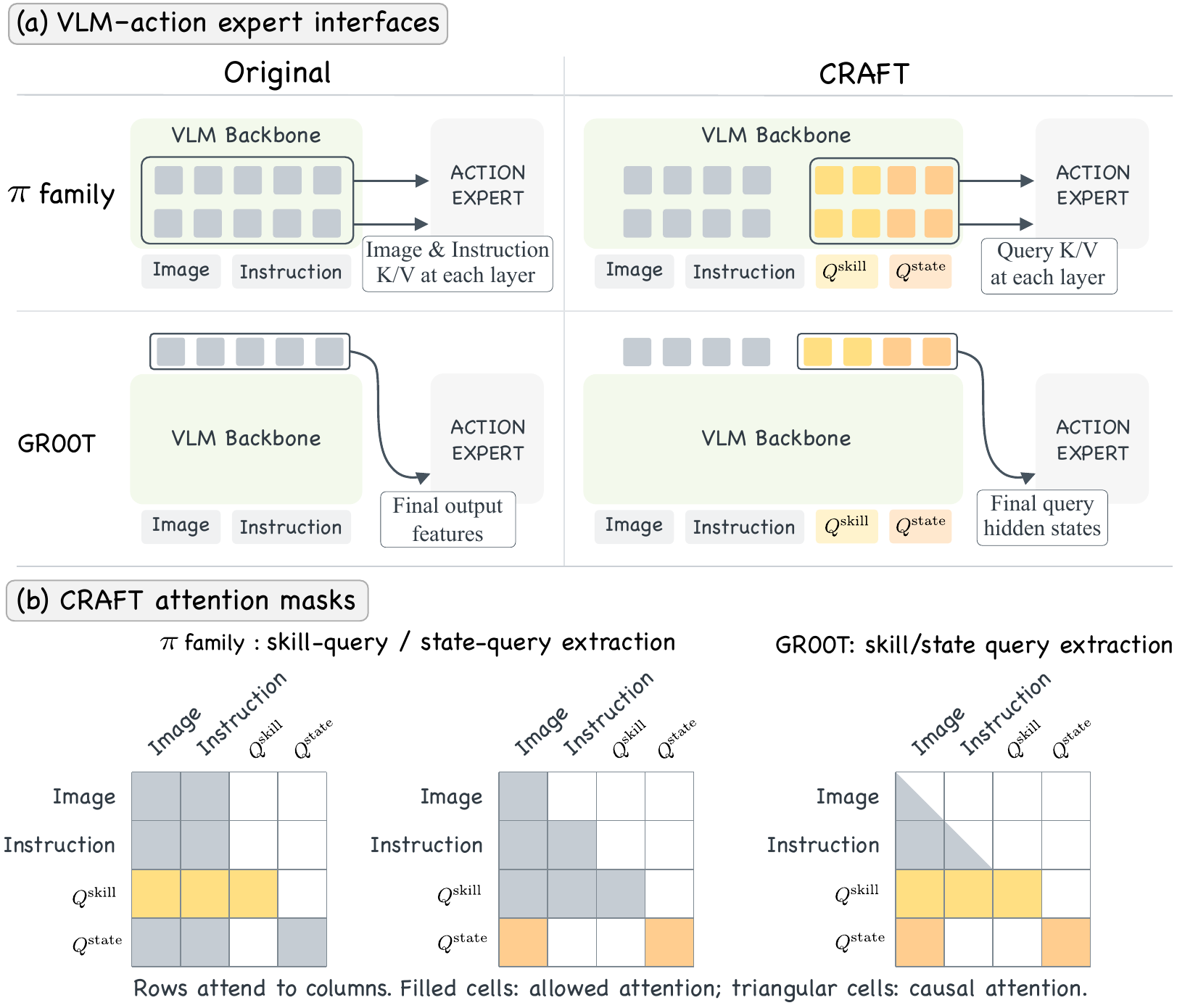}
\caption{
CRAFT architecture and attention masks.
(a) Original and CRAFT VLM--action expert interfaces.
CRAFT conditions the $\pi$-family action expert on layer-wise K/V from skill and state queries, and GR00T on their final hidden states.
(b) Attention masks used to construct the two representations.
Rows attend to columns; filled cells permit attention, white cells block it, and triangular cells denote causal attention.
The $\pi$ family extracts the two query representations in separate shared-backbone passes, whereas GR00T extracts both in a single pass.
Token counts are schematic; formatting and padding tokens are omitted.
}
\label{fig:craft-architecture}
\end{figure}

\paragraph{Query representations.}
CRAFT introduces learnable skill queries $Q^{\mathrm{skill}}$ and state queries $Q^{\mathrm{state}}$ to obtain the two representations described in Section~\ref{sec:method:rep}.
We use the same query-based representation design for the $\pi$ family ($\pi_0$ and $\pi_{0.5}$) and GR00T N1.7, while adapting its implementation to each model's VLM--action expert interface.
For the $\pi$ family, the representations are layer-wise K/V tensors at the query positions; for GR00T, they are the final hidden states at those positions.
Figure~\ref{fig:craft-architecture} compares the original and CRAFT interfaces and shows the corresponding attention masks.

\paragraph{The $\pi$ family.}
In the original $\pi$ family, the action expert attends to the keys and values (K/V) associated with image and instruction tokens at each VLM backbone layer.
In CRAFT, the expert instead attends to the K/V at the skill and state query positions, which constitute the skill and state representations, respectively.

\paragraph{GR00T.}
GR00T conditions its action expert on the final VLM features through cross-attention.
In CRAFT, this conditioning uses only the final hidden states at the skill and state query positions.

\paragraph{Attention structure.}
Skill queries attend to image and instruction tokens, whereas state queries attend to visual tokens without directly attending to instruction tokens.
Queries attend bidirectionally within their own group.
The skill and state query groups do not attend to each other.
Figure~\ref{fig:craft-architecture}(b) shows the attention masks used for each architecture.

\paragraph{Initialization.}
Each query token is initialized with the mean of the pretrained backbone's token embeddings plus independent Gaussian noise with standard deviation $0.02$.
The queries are learned during fine-tuning, with the same attention structure used at training and inference.

\subsection{Skill Annotation and Counterfactual Instructions}
\label{app:cf}

\paragraph{Skill annotation.}
CRAFT uses operation--entity labels to identify the skill executed at each timestep and to select samples across demonstrations according to skill.
In our tasks, the operation order is fixed and the instruction specifies the entity assigned to each operation.
Thus, the VLM does not infer skill identities; it only locates transitions between the known operations.
Following prior work on VLM-based demonstration segmentation~\citep{kuramshin2026tread,tabakov2026atomic}, we use a VLM to annotate demonstrations for both pick--place and pick--place--press tasks.

The VLM receives time-indexed demonstration frames, the instruction, and the
ordered skills, and returns the transition frame indices.
We assign the corresponding operation--entity label to each frame within a segment.
Figure~\ref{fig:boundary-prompt} shows the annotation prompt for pick--place demonstrations.
These labels are used only during training and are not provided to the policy at inference.

\begin{figure}[t]
\centering
\fbox{\begin{minipage}{\dimexpr\linewidth-2\fboxsep-2\fboxrule\relax}
\textbf{Demonstration:} \texttt{<time-indexed demonstration frames>}\par
\textbf{Instruction:} Pick the blue cube and place it on the blue plate.\par
\textbf{Ordered skills:} pick blue cube $\rightarrow$ place blue plate.
\medskip

Locate the first frame in which the grasped cube begins moving with the gripper, counting the initial lift as part of placing.
Use the following frames to check that the motion continues.
Return the frame ID and note any uncertainty.
\end{minipage}}
\caption{
VLM prompt for pick--place skill annotation.
The transition is defined as the first frame in which the grasped object begins moving with the gripper, with the initial lift assigned to placing.
}

\label{fig:boundary-prompt}
\end{figure}

\paragraph{Constructing counterfactual instructions.}
Given the annotated current operation, we construct counterfactual instructions by changing one or more entity assignments for the current or future operations while keeping the observation and operation order fixed.
Assignments for completed operations remain unchanged because their outcomes are already reflected in the observation.
For each edited operation, we select a replacement entity that has been used for that operation in the demonstrations.
We exclude combinations of entity assignments already present in the demonstration set.

\paragraph{Changing or preserving the current skill.}
We distinguish two cases according to whether the counterfactual instruction changes the currently required skill.
A \emph{skill-changing} counterfactual changes the entity assigned to the current operation and may also change future assignments.
A \emph{skill-preserving} counterfactual keeps the current-operation entity unchanged and changes one or more future assignments.
Table~\ref{tab:cf-construction} illustrates both cases during placing in a pick--place--press task.
The counterfactual type determines the reference sample used for supervision (Appendix~\ref{app:train:align}).

\begin{table}[htbp]
\centering
\caption{
Counterfactual instructions during placing in the pick--place--press task, whose demonstrations contain only all-blue and all-red combinations.
Bold entries indicate changed entity assignments.
Both counterfactuals retain the completed pick assignment and specify an undemonstrated combination.
}
\label{tab:cf-construction}
\begin{tabular}{llll}
\toprule
Variant & Pick (completed) & Place (current) & Press (future) \\
\midrule
Original & Blue cube & Blue plate & Blue button \\
Skill-changing & Blue cube & \textbf{Red plate} & Blue button \\
Skill-preserving & Blue cube & Blue plate & \textbf{Red button} \\
\bottomrule
\end{tabular}
\end{table}

\subsection{Training}
\label{app:train}

We detail the demonstrated-skill comparisons, counterfactual supervision, and joint optimization used by CRAFT.

\subsubsection{Learning Skill Representations from Demonstrations}
\label{app:train:rep}

\paragraph{Selecting comparison samples.}
We form batches with multiple samples per skill using the operation--entity labels from Appendix~\ref{app:cf}.
For each sample, we select a positive from another execution with the same label and negatives that execute the same operation on different entities.
This focuses the comparison on entity differences within the same operation.

\paragraph{Controlling the comparison.}
For all comparisons involving a given sample, we hold its state representation, noisy action chunk, flow time, and target fixed.
We vary only the skill representation, using the selected positive and negative samples, and evaluate the resulting predictions with the score in Equation~\ref{eq:skill-score}.

\paragraph{Normalizing the prediction error.}
To account for differences in target magnitude across samples, we normalize the squared prediction error by the expected squared magnitude of the target velocity.
Let $D_i$ denote the number of action elements used to compute the score.
For a fixed demonstrated action chunk and independent standard Gaussian noise, this scale is
\begin{equation}
\label{eq:score-scale}
\sigma_i
=
\mathbb{E}_{\epsilon_i}\!\left[
\|u_i\|_2^2
\mid \mathbf{a}_i
\right]
=
D_i\left(1+\operatorname{mean}(\mathbf{a}_i^2)\right).
\end{equation}
The term $1$ corresponds to the unit variance of the Gaussian noise, and the second term is the mean squared action magnitude.
In implementation, we equivalently divide the mean squared prediction error by
$1+\operatorname{mean}(\mathbf{a}_i^2)$.
The prediction error and normalization scale are computed over the same action dimensions and timesteps.

\subsubsection{Counterfactual Skill-Representation Alignment}
\label{app:train:align}

\paragraph{Matching counterfactuals to demonstrations.}
Within each batch, we match skill-changing counterfactuals with demonstrated reference samples sharing the required operation--entity label.
The counterfactual supplies the skill representation, while the reference supplies its state representation, noisy action chunk, flow time, and demonstrated velocity target.
For skill-preserving supervision, the reference is the original sample itself.
Both cases use the predictions and targets in Equations~\ref{eq:cf-pred}--\ref{eq:cf}.

\paragraph{Supervising the current operation.}
The counterfactual and reference are matched by the currently required skill, while their subsequent operations may differ.
Because an action chunk can cross operation boundaries, we restrict each counterfactual loss to the timesteps of the reference action chunk that remain within the matched operation.
The demonstrated flow-matching loss continues to supervise the complete valid action chunk.

\paragraph{Aggregating the losses.}
We compute the changing and preserving terms separately, averaging squared prediction error over action dimensions and the retained timesteps, and add them as in Equation~\ref{eq:cf}.
The preserving term applies only when a future entity assignment can be changed.

\subsubsection{Joint Optimization}
\label{app:train:obj}

\paragraph{Assigning updates to each loss.}
The skill objective jointly learns reusable skill representations and the action expert that interprets them.
For counterfactual alignment, we hold the reference state representation and action expert parameters fixed for this loss, training the counterfactual skill representation to elicit the required behavior through the fixed expert.
Table~\ref{tab:gradient-routing} summarizes the gradient paths.

\begin{table}[htbp]
\centering
\caption{
Gradient paths for the training objectives.
Each row describes the contribution of one loss to the joint update.
}
\label{tab:gradient-routing}
\begin{tabular}{llll}
\toprule
Loss & Skill representation & State representation & Action expert \\
\midrule
$\mathcal L_{\mathrm{fm}}$
& Gradient passes & Gradient passes & Updated \\
$\mathcal L_{\mathrm{skill}}$
& Gradient passes & Gradient stopped & Updated \\
$\mathcal L_{\mathrm{cf}}$
& Gradient passes & Gradient stopped & Fixed for this loss \\
\bottomrule
\end{tabular}
\end{table}

Stopping gradients through the state representation does not freeze the shared backbone, which remains trainable through the skill pathway and the demonstrated flow-matching loss.

\paragraph{Joint training.}
We optimize the pretrained backbone, action expert, and queries jointly in a single training stage using Equation~\ref{eq:total}, retaining each model's action normalization and flow-matching convention.
We gradually introduce the auxiliary objectives by increasing $\beta$ and $\gamma$ from zero to their target values during warmup, while keeping the demonstrated flow-matching loss active throughout.
Both auxiliary weights remain fixed after warmup.
Skill labels, comparison samples, and counterfactual instructions are used only during training.

\subsection{Implementation Details}
\label{app:implementation}

\paragraph{Training configuration.}
For the simulation experiments, we train each method for 20k optimization steps using NVIDIA H200 GPUs.
For CRAFT, we construct skill-balanced batches with four samples per skill on \textsc{Pick-Place} and three samples per skill on \textsc{Pick-Place-Press}, giving batch sizes of 32 and 18, respectively.
CRAFT uses three negative samples per anchor, a contrastive temperature of $\tau=0.1$, and loss weights $\beta=0.03$ and $\gamma=1.0$, with equal weight on the skill-changing and skill-preserving terms.
The auxiliary-loss weights increase from zero to their target values over the first 3k steps and remain fixed thereafter.
We use 64 queries in total: 48 skill and 16 state queries for $\pi_{0.5}$ on \textsc{Pick-Place}, and 32 of each for all other model--benchmark configurations.

\paragraph{Optimization and baselines.}
We use AdamW with a 1k-step learning-rate warmup followed by cosine decay.
The peak learning rate is $2.5\times10^{-5}$ for $\pi_0$ and $\pi_{0.5}$, and $5\times10^{-5}$ for GR00T.
For the $\pi$ models, we clip the gradient norm at $1.0$ and use an EMA decay of $0.999$.
The LoRA baselines use rank $16$, with $\alpha=16$ for the $\pi$ models and $\alpha=32$ for GR00T.
For Skill Composition, we use the same skill annotations as CRAFT to form constituent-skill training segments and a rule based on the observed gripper state to advance between skills at deployment.

\paragraph{Checkpoint selection.}
Following prior VLA evaluations that periodically evaluate training checkpoints
and report the best-performing checkpoint~\citep{kim2025oft, li2026cogvla}, we evaluate checkpoints every 2k steps from 10k to 20k.
For each method, model, and benchmark, we select the checkpoint with the highest
mean success on undemonstrated combinations across the three evaluation seeds,
using these evaluation rollouts for both selection and reporting.

\paragraph{Deployment.}
Policies predict 20-step action chunks and execute the first 10 actions before replanning.
We retain each model's default number of flow-integration steps: 10 for $\pi_0$ and $\pi_{0.5}$, and 4 for GR00T.

\section{Benchmark Construction}
\label{app:benchmark}

\begin{figure}[t]
\centering
\includegraphics[width=\linewidth]{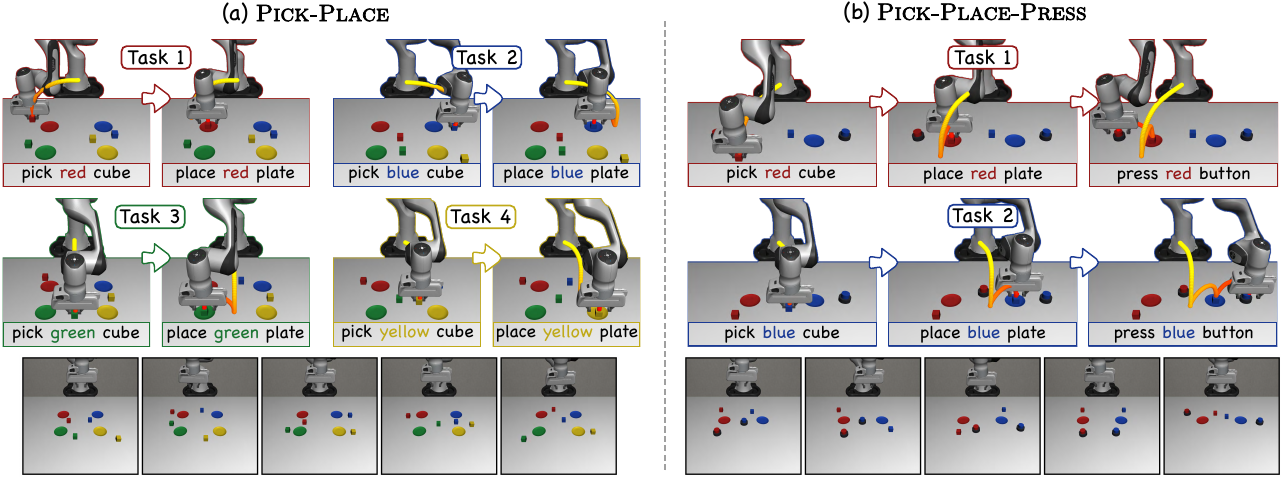}
\caption{
Benchmark construction for (a) \textsc{Pick-Place} and (b) \textsc{Pick-Place-Press}.
Top: demonstrated same-color combinations used for data collection.
Bottom: example initial scenes from each benchmark.
The remaining combinations form the undemonstrated evaluation set.
}
\label{fig:benchmark-details}
\end{figure}

We construct two tabletop benchmarks in LIBERO~\citep{liu2023libero} to evaluate compositional generalization to undemonstrated combinations of demonstrated skills.
In both benchmarks, the operation sequence is fixed, and each instruction specifies the entity assigned to each operation.
Scenes contain the same set of objects across task combinations, while object positions vary across episodes.
Demonstrations cover only same-color combinations, shown at the top of Figure~\ref{fig:benchmark-details}; the remaining combinations require recombining demonstrated operation--entity skills.

For each benchmark, we generate a shared set of 50 initial scenes and collect one demonstration from each scene for every demonstrated combination using SpaceMouse teleoperation.
Evaluation uses a separate set of initial scenes sampled from the same initialization distribution.
We allow up to 300 environment steps per rollout for \textsc{Pick-Place} and 500 for \textsc{Pick-Place-Press}, which contains one additional operation.

\subsection{\textsc{Pick-Place}}
\label{app:benchmark:pick-place}

The scene contains one cube and one plate in each of four colors: red, blue, green, and yellow.
Each instruction specifies which cube to pick and which plate to place it on.
The four same-color combinations cover all eight constituent skills---four picking skills and four placing skills---while leaving the other 12 of the 16 combinations undemonstrated.

The plates occupy fixed positions in a four-quadrant layout.
Each cube is initialized at a sampled position within the quadrant containing its same-color plate, subject to spacing constraints that separate the objects.
This produces variation in the cube positions and corresponding manipulation trajectories across initial scenes.
Task success requires the specified cube to be on the specified plate, as determined by a predicate based on contact and relative position.

\subsection{\textsc{Pick-Place-Press}}
\label{app:benchmark:pick-place-press}

This benchmark extends the operation sequence to picking a cube, placing it on a plate, and pressing a button.
The scene contains one red and one blue instance of each object type.
Demonstrations cover the all-red and all-blue combinations, providing all six constituent skills while leaving the other six of the eight combinations undemonstrated.

The two plates occupy fixed positions, while the cube and button of each color are initialized around the corresponding plate.
Task success requires the specified cube to be on the specified plate and the end-effector to be within a horizontal tolerance of the specified button center and below a height threshold.

\section{Additional Results}
\label{app:results}

\subsection{Performance across Task Combinations}
\label{app:results:combinations}

\begin{figure}[t]
\centering
\includegraphics[width=0.9\linewidth]{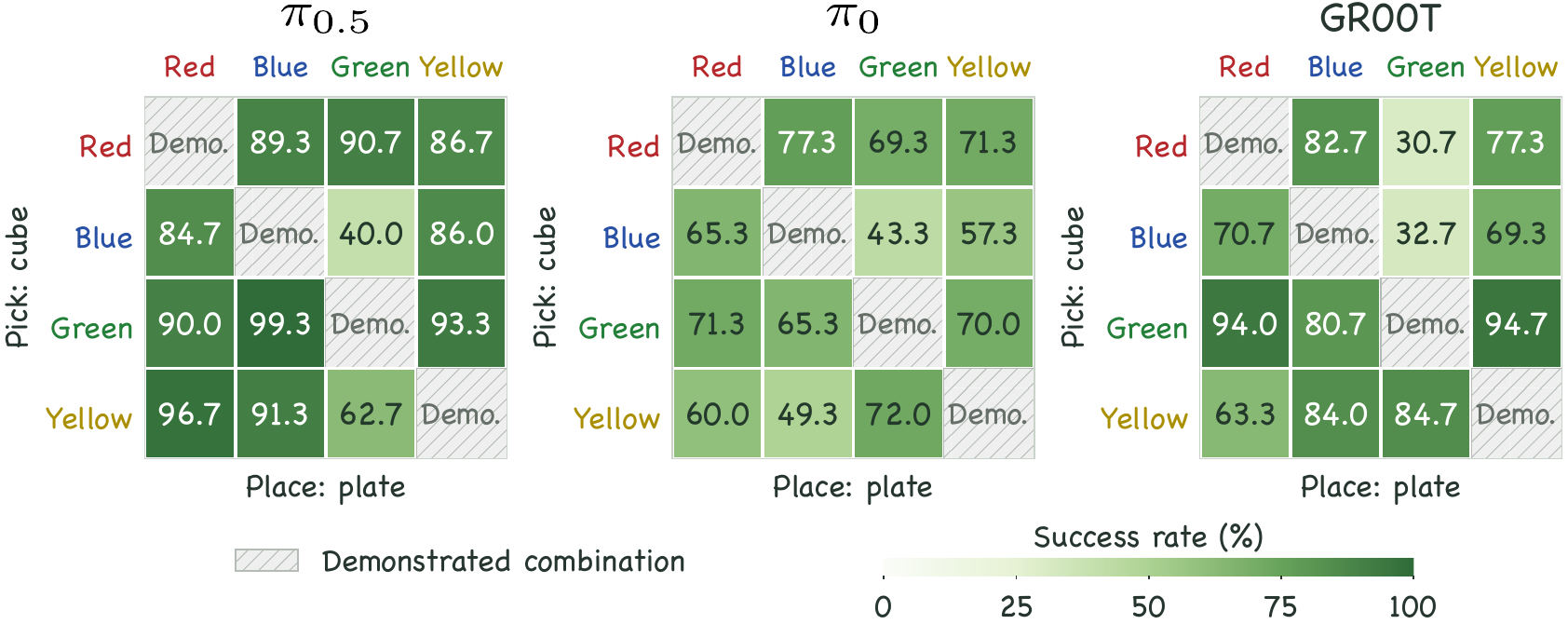}
\caption{
CRAFT success rates (\%) across \textsc{Pick-Place} task combinations, averaged over three evaluation seeds.
Rows and columns specify cube and plate colors, respectively.
Hatched cells denote demonstrated combinations; all panels use the same color scale.
}
\label{fig:results:pick-place}
\end{figure}

\begin{figure}[t]
\centering
\includegraphics[width=0.9\linewidth]{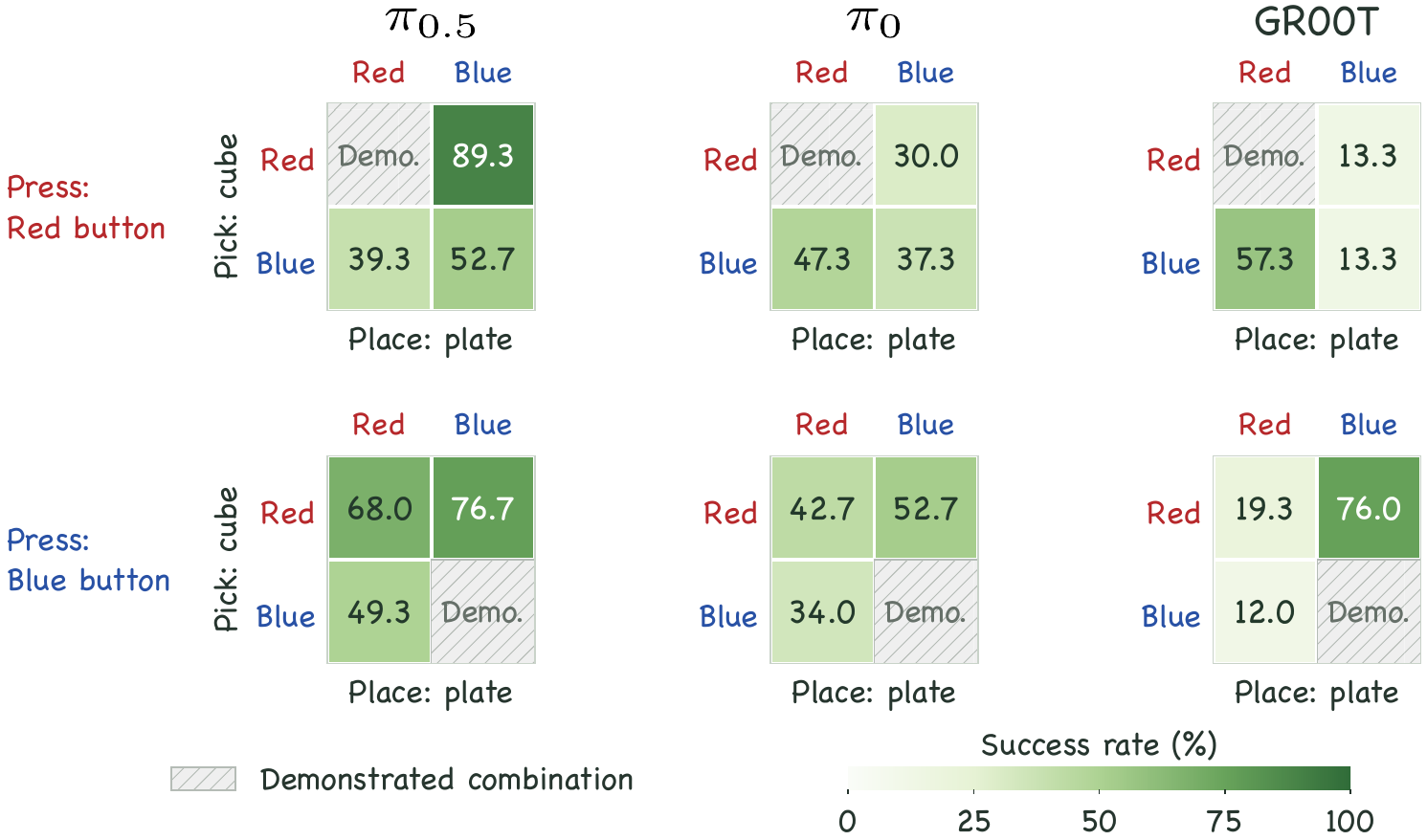}
\caption{
CRAFT success rates (\%) across \textsc{Pick-Place-Press} task combinations, averaged over three evaluation seeds.
Rows and columns within each matrix specify cube and plate colors, respectively; the upper and lower matrices correspond to red and blue buttons.
Hatched cells denote demonstrated combinations; all panels use the same color scale.
}
\label{fig:results:pick-place-press}
\end{figure}

We break down CRAFT's success across individual undemonstrated combinations to examine variation within each benchmark.
Figures~\ref{fig:results:pick-place} and~\ref{fig:results:pick-place-press} report success for each task combination, averaged over three evaluation seeds with 50 trials per combination and seed.
Hatched cells identify demonstrated combinations, whose aggregate success rates are reported in Table~\ref{tab:main}.

On \textsc{Pick-Place}, success varies substantially across combinations even though every constituent skill is demonstrated.
The blue-cube/green-plate combination is among the lowest-success combinations for all three models.
Other patterns are model-dependent: the red-cube/green-plate combination reaches $90.7\%$ success for $\pi_{0.5}$ but only $30.7\%$ for GR00T N1.7.
Thus, the combinations that remain difficult after CRAFT differ across VLA models.

On \textsc{Pick-Place-Press}, GR00T N1.7's success is concentrated in two combinations: red--blue--blue and blue--red--red, where the colors specify the cube, plate, and button, respectively.
The other four undemonstrated combinations have substantially lower success.
In contrast, $\pi_{0.5}$ distributes its success more broadly across the six undemonstrated combinations.

\subsection{Action Predictions under Instruction Changes}
\label{app:results:instruction-response}
\begin{figure}[t]
\centering
\includegraphics[width=0.9\linewidth]{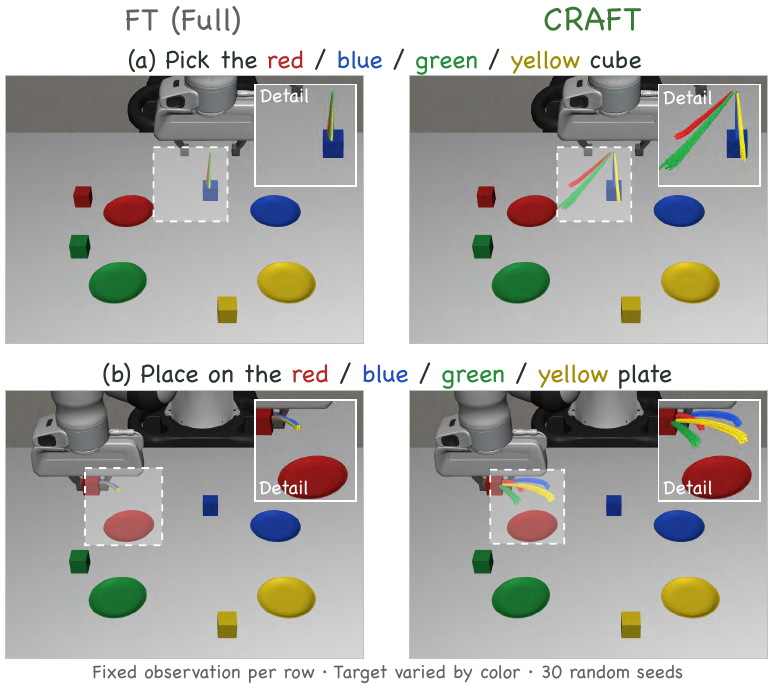}
\caption{
Action predictions under instruction changes with a fixed observation.
Top: the observation is taken while picking the blue cube, and the instructed cube is varied across four colors.
Bottom: the observation is taken while placing on the red plate, and the instructed plate is varied across four colors.
The blue target in the top row and red target in the bottom row correspond to the original executions.
Trajectory colors indicate the instructed entity, with predicted action chunks shown for 30 random seeds.
Insets enlarge the marked regions at the same scale within each row.
}
\label{fig:instruction-response}
\end{figure}

Using $\pi_{0.5}$ on \textsc{Pick-Place}, we examine how action predictions respond to instruction changes while holding the visual observation fixed.
For the picking example, we fix an observation from an execution of picking the blue cube and vary the instructed cube among the four colors.
For the placing example, we fix an observation from an execution of placing on the red plate and similarly vary the instructed plate.
Thus, the blue instruction in the first row and the red instruction in the second correspond to the original execution, while the other colors change the currently required skill without changing the observation.
For each instruction, we visualize predicted action chunks from 30 random seeds (Figure~\ref{fig:instruction-response}).

Under CRAFT, the predicted trajectories separate according to the instructed cube or plate.
In contrast, predictions from FT (\textit{Full}) largely overlap across instruction changes in both examples.
This qualitative difference suggests that CRAFT's action predictions are more responsive to the currently required skill under a fixed observation.

\subsection{Reusing Skill Representations across Executions}
\label{app:results:skill-swap}

We test whether CRAFT's skill representations can be reused across executions using $\pi_{0.5}$ on \textsc{Pick-Place}.
For each recipient observation, we replace its skill-query K/V with those from another execution while keeping the recipient's state representation and sampling noise fixed.
Donors execute the same operation as the recipient, with either the same or a different entity.
We evaluate 120 recipient frames, with 15 per skill, and sample 25 donors for each candidate donor skill.
For each substitution, we measure the relative change in the predicted action chunk, normalized by the norm of the original prediction.

\begin{figure}[t]
\centering
\includegraphics[width=0.8\linewidth]{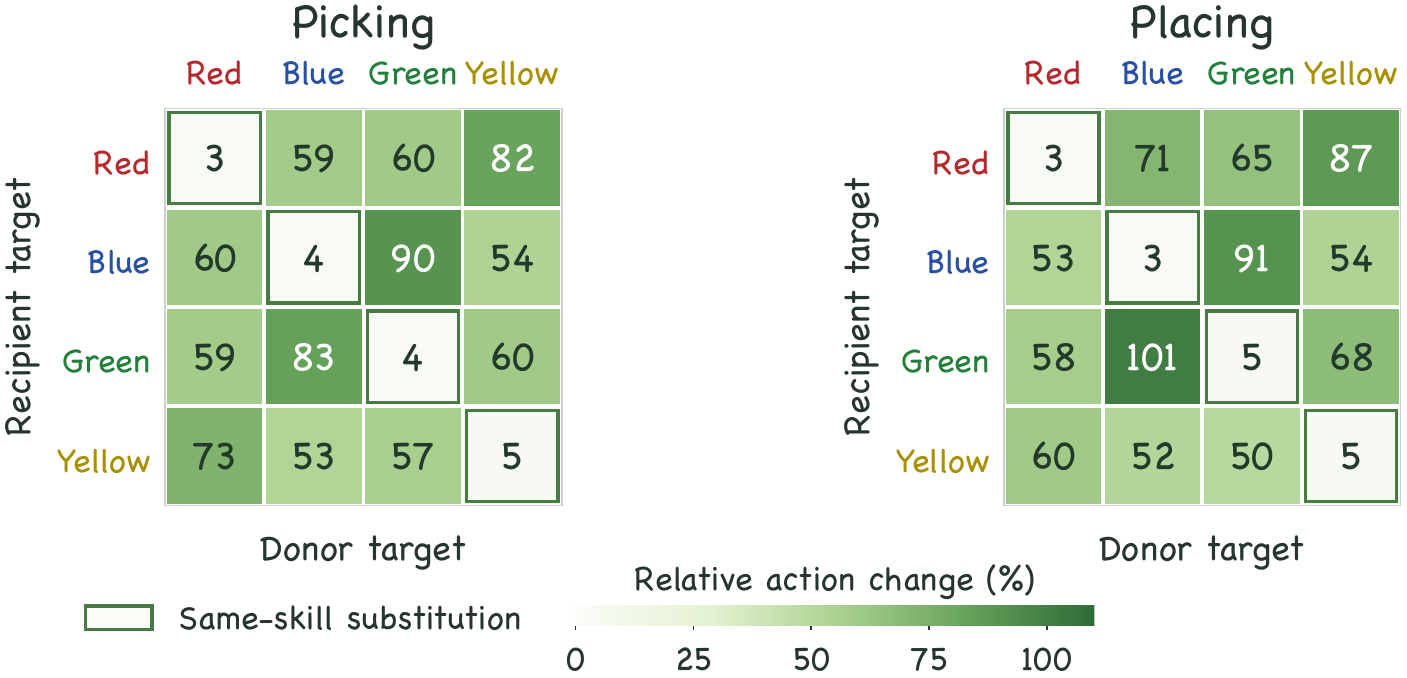}
\caption{
Skill-representation reuse across executions on \textsc{Pick-Place} with $\pi_{0.5}$.
For each recipient observation, skill-query K/V are replaced with those from another execution while the recipient's state representation and sampling noise remain fixed.
Rows and columns indicate recipient and donor entities, respectively.
Each cell reports the mean relative change (\%) in the predicted action chunk.
Outlined diagonal cells use donors executing the same skill as the recipient; off-diagonal cells use donors executing the same operation with a different entity.
Both panels use the same color scale.
}
\label{fig:skill-swap}
\end{figure}

Figure~\ref{fig:skill-swap} shows a clear contrast between same-skill and different-skill substitutions for both picking and placing.
Replacing the skill representation with one from another execution of the same skill changes the prediction by only $4.0\pm3.1\%$ on average across recipients.
In contrast, substitutions from different skills produce substantially larger changes across donor--recipient pairs.
These results support the intended reuse of skill representations across executions: representations from different executions of the same skill yield similar action predictions under a fixed recipient state, while representations of different skills produce distinct predictions.

\subsection{Query Count and Allocation}
\label{app:results:queries}

\begin{figure}[t]
\centering
\includegraphics[width=\linewidth]{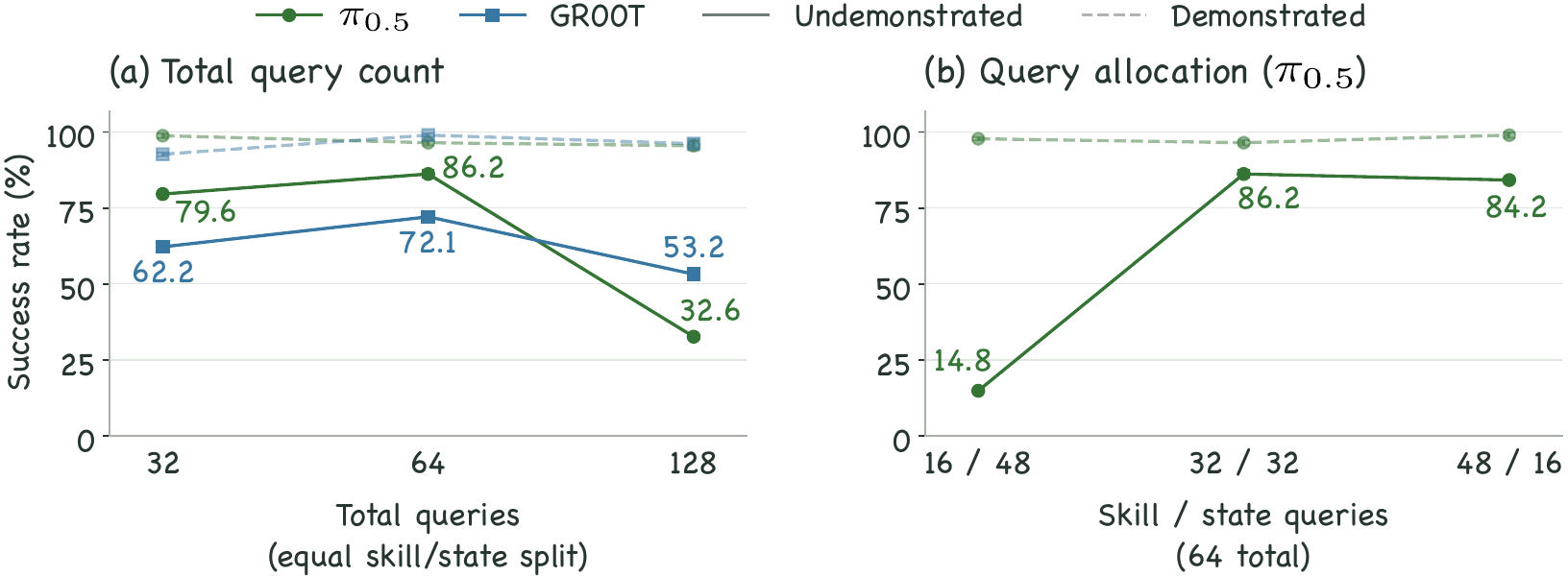}
\caption{
Query-count and allocation ablations on \textsc{Pick-Place}.
(a) Total query count for $\pi_{0.5}$ and GR00T N1.7, with an equal number of skill and state queries.
(b) Skill/state query allocation for $\pi_{0.5}$ with 64 total queries.
Solid and dashed lines denote success on undemonstrated and demonstrated combinations, respectively.
}
\label{fig:results:queries}
\end{figure}

We examine the sensitivity of compositional generalization to the number and allocation of skill and state queries on \textsc{Pick-Place} (Figure~\ref{fig:results:queries}).
For the query-count sweep, skill and state queries are allocated equally.
For the allocation sweep, we fix the total at 64 queries and vary the skill/state split for $\pi_{0.5}$.

\paragraph{Total query count.}
For both $\pi_{0.5}$ and GR00T N1.7, success on undemonstrated combinations increases from 32 to 64 total queries and decreases at 128.
Thus, increasing the number of queries does not monotonically improve compositional generalization.
Success on demonstrated combinations remains high across the sweep, indicating that the degradation at 128 queries is concentrated on undemonstrated combinations.

\paragraph{Skill/state allocation.}
With 64 total queries, the state-heavy 16/48 skill/state allocation yields substantially lower success on undemonstrated combinations than the 32/32 and 48/16 allocations, while demonstrated success remains high across all three settings.
Within this sweep, performance is therefore lower when a larger share of the query budget is allocated to the state representation than to the skill representation.
The balanced 32/32 and skill-heavy 48/16 allocations perform similarly well.
The main $\pi_{0.5}$ configuration uses 48/16 skill/state queries, while the main GR00T N1.7 configuration uses 32/32.

\subsection{Guidance-Strength Selection for CAG-VA}
\label{app:results:cagva}

\paragraph{CAG-VA implementation.}
We implement the vision-action (VA) variant of CAG-VA~\citep{fang2026vision}.
For each backbone and adaptation setting, we fine-tune a separate VA policy on the same demonstrations while masking the language input.
At inference, CAG-VA combines the language-conditioned VLA policy
$\pi_{\mathrm{cond}}(a\mid o,\ell)$ with the corresponding vision-only policy
$\pi_{\mathrm{VA}}(a\mid o)$ using
\begin{equation}
\pi_{\mathrm{CAG}}(a\mid o,\ell)
=
\pi_{\mathrm{VA}}(a\mid o)
+
\omega
\left(
\pi_{\mathrm{cond}}(a\mid o,\ell)
-
\pi_{\mathrm{VA}}(a\mid o)
\right),
\end{equation}
where $\omega$ controls the guidance strength.
For the flow-based VLAs, we apply this guidance rule to the velocity predictions at each flow-integration step.
Thus, $\omega=1$ recovers the original language-conditioned VLA, while larger values amplify the difference between the language-conditioned and vision-only predictions.
Both branches are used at every inference step.

\begin{figure}[t]
\centering
\includegraphics[width=0.85\linewidth]{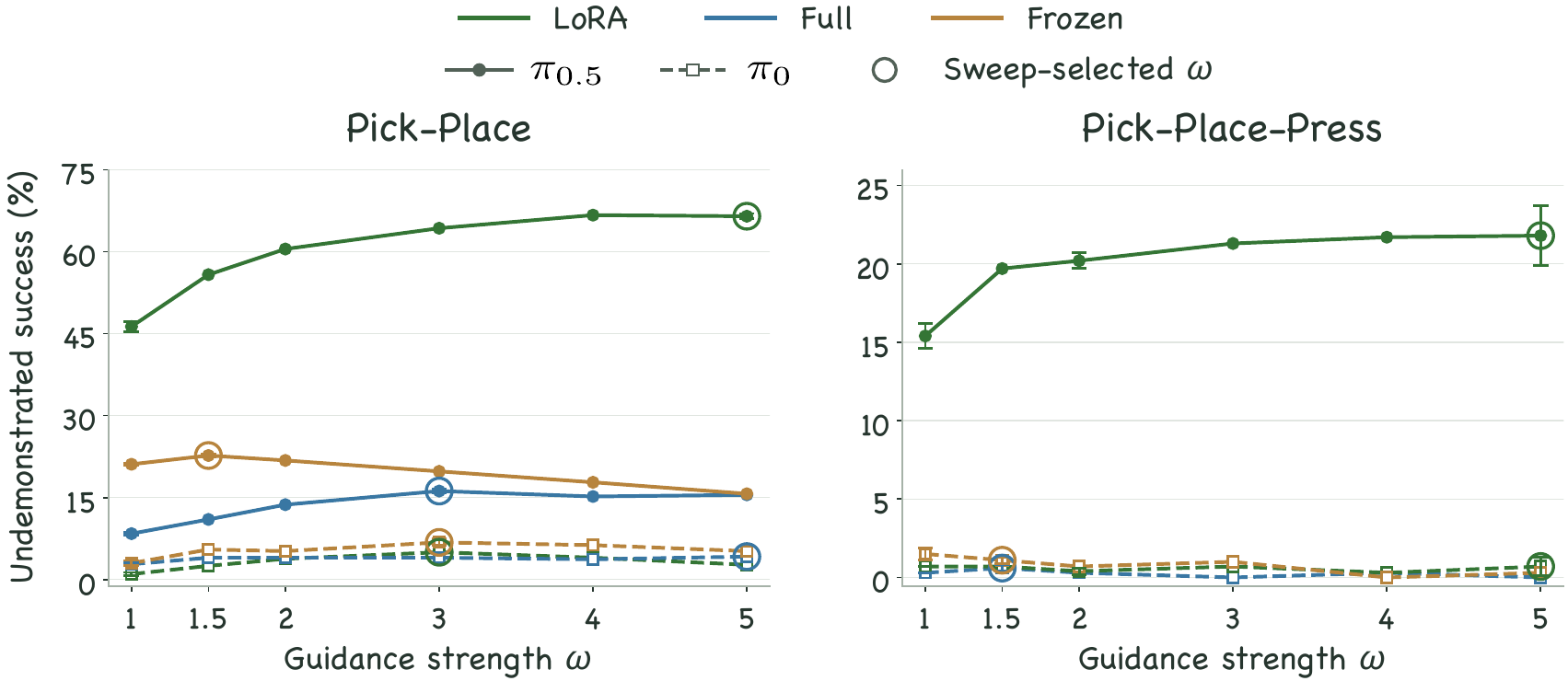}
\caption{
CAG-VA guidance-strength sweeps on \textsc{Pick-Place} and
\textsc{Pick-Place-Press}.
Each configuration uses a separately fine-tuned vision-only VA policy together with the corresponding language-conditioned VLA policy.
Curves show success on undemonstrated combinations as a function of guidance strength $\omega$.
The initial sweep uses seed 7; selected settings additionally evaluated with seeds 0 and 42 are shown as three-seed means with standard-deviation error bars, while the remaining points show seed-7 results.
Outlined markers denote the best guided setting ($\omega>1$) in each seed-7 sweep; $\omega=1$ recovers the unguided language-conditioned policy.
}
\label{fig:cagva-omega}
\end{figure}

\paragraph{Guidance-strength selection.}
We evaluate $\omega \in \{1, 1.5, 2, 3, 4, 5\}$ separately for each
model--adaptation--benchmark configuration, treating $\omega = 1$ as the
unguided reference.
For each model and benchmark, we use seed 7 to select the guided configuration
($\omega > 1$) with the highest undemonstrated-combination success across
adaptation settings and guidance strengths, and evaluate that configuration
additionally with seeds 0 and 42.
Figure~\ref{fig:cagva-omega} and Table~\ref{tab:cagva-omega} show the sweeps
for the $\pi$ models.
Points evaluated with all three seeds are reported as three-seed means, whereas
all remaining points show the seed-7 sweep results.
For GR00T N1.7, we follow the same selection procedure; the selected
configurations are Frozen with $\omega = 1.5$ on \textsc{Pick-Place} and LoRA
with $\omega = 3$ on \textsc{Pick-Place-Press}.
The main results report three-seed performance for the selected CAG-VA
configuration for each model and benchmark.

\paragraph{Effect of guidance strength.}
The effect of guidance varies substantially across models and adaptation settings.
For $\pi_{0.5}$ with LoRA, stronger guidance improves success on undemonstrated combinations on both benchmarks, although high guidance strengths also reduce success on demonstrated combinations.
For $\pi_0$, the gains are considerably smaller, and stronger guidance can reduce demonstrated success without a corresponding improvement on undemonstrated combinations.
These results motivate selecting $\omega$ separately for each configuration rather than using a single guidance strength across models and adaptation settings.

\begin{table}[t]
\centering
\caption{
CAG-VA guidance-strength sweep.
Each cell reports undemonstrated success with demonstrated success in parentheses (\%).
The initial sweep uses seed 7; selected settings additionally evaluated with seeds 0 and 42 are reported using their three-seed means.
Bold entries indicate the final CAG-VA configurations reported in Table~\ref{tab:main} and include mean $\pm$ standard deviation over the three evaluation seeds.
The setting $\omega=1$ corresponds to the unguided language-conditioned policy.
}
\label{tab:cagva-omega}
\small
\setlength{\tabcolsep}{3pt}
\renewcommand{\arraystretch}{1.1}
\begin{tabular*}{\linewidth}{@{\extracolsep{\fill}}lcccccc@{}}
\toprule
Configuration & $\omega=1$ & $1.5$ & $2$ & $3$ & $4$ & $5$ \\
\midrule
\multicolumn{7}{@{}l}{\textsc{Pick-Place}} \\
\addlinespace[2pt]
\raisebox{0.5\baselineskip}{$\pi_{0.5}$ LoRA} & \shortstack[c]{$46.3$\\$(100)$} & \shortstack[c]{$55.8$\\$(100)$} & \shortstack[c]{$60.5$\\$(100)$} & \shortstack[c]{$64.3$\\$(100)$} & \shortstack[c]{$65.7$\\$(99.5)$} & \shortstack[c]{$\mathbf{66.5\pm 0.5}$\\$(\mathbf{96.3\pm 1.5})$} \\
\addlinespace[3pt]
\raisebox{0.5\baselineskip}{$\pi_{0.5}$ Full} & \shortstack[c]{$8.4$\\$(100)$} & \shortstack[c]{$11.0$\\$(100)$} & \shortstack[c]{$13.7$\\$(99.5)$} & \shortstack[c]{$16.2$\\$(98.5)$} & \shortstack[c]{$15.2$\\$(95)$} & \shortstack[c]{$15.5$\\$(94.5)$} \\
\addlinespace[3pt]
\raisebox{0.5\baselineskip}{$\pi_{0.5}$ Frozen} & \shortstack[c]{$21.1$\\$(99)$} & \shortstack[c]{$22.7$\\$(98.7)$} & \shortstack[c]{$21.8$\\$(97.5)$} & \shortstack[c]{$19.8$\\$(95.5)$} & \shortstack[c]{$17.8$\\$(86)$} & \shortstack[c]{$15.7$\\$(83)$} \\
\addlinespace[3pt]
\cmidrule(lr){1-7}
\raisebox{0.5\baselineskip}{$\pi_0$ LoRA} & \shortstack[c]{$1.0$\\$(97.5)$} & \shortstack[c]{$2.5$\\$(96.5)$} & \shortstack[c]{$3.8$\\$(93)$} & \shortstack[c]{$5.0$\\$(81)$} & \shortstack[c]{$4.0$\\$(71)$} & \shortstack[c]{$2.7$\\$(57)$} \\
\addlinespace[3pt]
\raisebox{0.5\baselineskip}{$\pi_0$ Full} & \shortstack[c]{$2.8$\\$(100)$} & \shortstack[c]{$4.0$\\$(100)$} & \shortstack[c]{$4.0$\\$(100)$} & \shortstack[c]{$4.0$\\$(97.5)$} & \shortstack[c]{$3.7$\\$(93)$} & \shortstack[c]{$4.2$\\$(87)$} \\
\addlinespace[3pt]
\raisebox{0.5\baselineskip}{$\pi_0$ Frozen} & \shortstack[c]{$3.0$\\$(48.2)$} & \shortstack[c]{$5.5$\\$(57)$} & \shortstack[c]{$5.2$\\$(53.5)$} & \shortstack[c]{$\mathbf{6.8\pm 0.6}$\\$(\mathbf{50.5\pm 1.3})$} & \shortstack[c]{$6.3$\\$(40)$} & \shortstack[c]{$5.2$\\$(36.5)$} \\
\addlinespace[3pt]
\midrule
\multicolumn{7}{@{}l}{\textsc{Pick-Place-Press}} \\
\addlinespace[2pt]
\raisebox{0.5\baselineskip}{$\pi_{0.5}$ LoRA} & \shortstack[c]{$15.4$\\$(100)$} & \shortstack[c]{$19.7$\\$(100)$} & \shortstack[c]{$20.2$\\$(99)$} & \shortstack[c]{$21.3$\\$(96)$} & \shortstack[c]{$21.7$\\$(94)$} & \shortstack[c]{$\mathbf{21.8\pm 1.9}$\\$(\mathbf{92.7\pm 3.5})$} \\
\addlinespace[3pt]
\cmidrule(lr){1-7}
\raisebox{0.5\baselineskip}{$\pi_0$ LoRA} & \shortstack[c]{$0.7$\\$(99.3)$} & \shortstack[c]{$0.7$\\$(96)$} & \shortstack[c]{$0.4$\\$(97.7)$} & \shortstack[c]{$0.7$\\$(87)$} & \shortstack[c]{$0.3$\\$(88)$} & \shortstack[c]{$0.7$\\$(81)$} \\
\addlinespace[3pt]
\raisebox{0.5\baselineskip}{$\pi_0$ Full} & \shortstack[c]{$0.3$\\$(99.7)$} & \shortstack[c]{$0.6$\\$(98.3)$} & \shortstack[c]{$0.3$\\$(97)$} & \shortstack[c]{$0.0$\\$(95)$} & \shortstack[c]{$0.3$\\$(90)$} & \shortstack[c]{$0.0$\\$(91)$} \\
\addlinespace[3pt]
\raisebox{0.5\baselineskip}{$\pi_0$ Frozen} & \shortstack[c]{$1.5$\\$(82)$} & \shortstack[c]{$\mathbf{1.1\pm 0.2}$\\$(\mathbf{69\pm 2.6})$} & \shortstack[c]{$0.7$\\$(64)$} & \shortstack[c]{$1.0$\\$(40)$} & \shortstack[c]{$0.0$\\$(24)$} & \shortstack[c]{$0.3$\\$(6)$} \\
\addlinespace[3pt]
\bottomrule
\end{tabular*}
\end{table}

\vspace{5cm}
\subsection{Additional Real-Robot Rollouts}
\label{app:results:real-rollouts}

\begin{figure}[h]
\centering
\includegraphics[width=\linewidth]{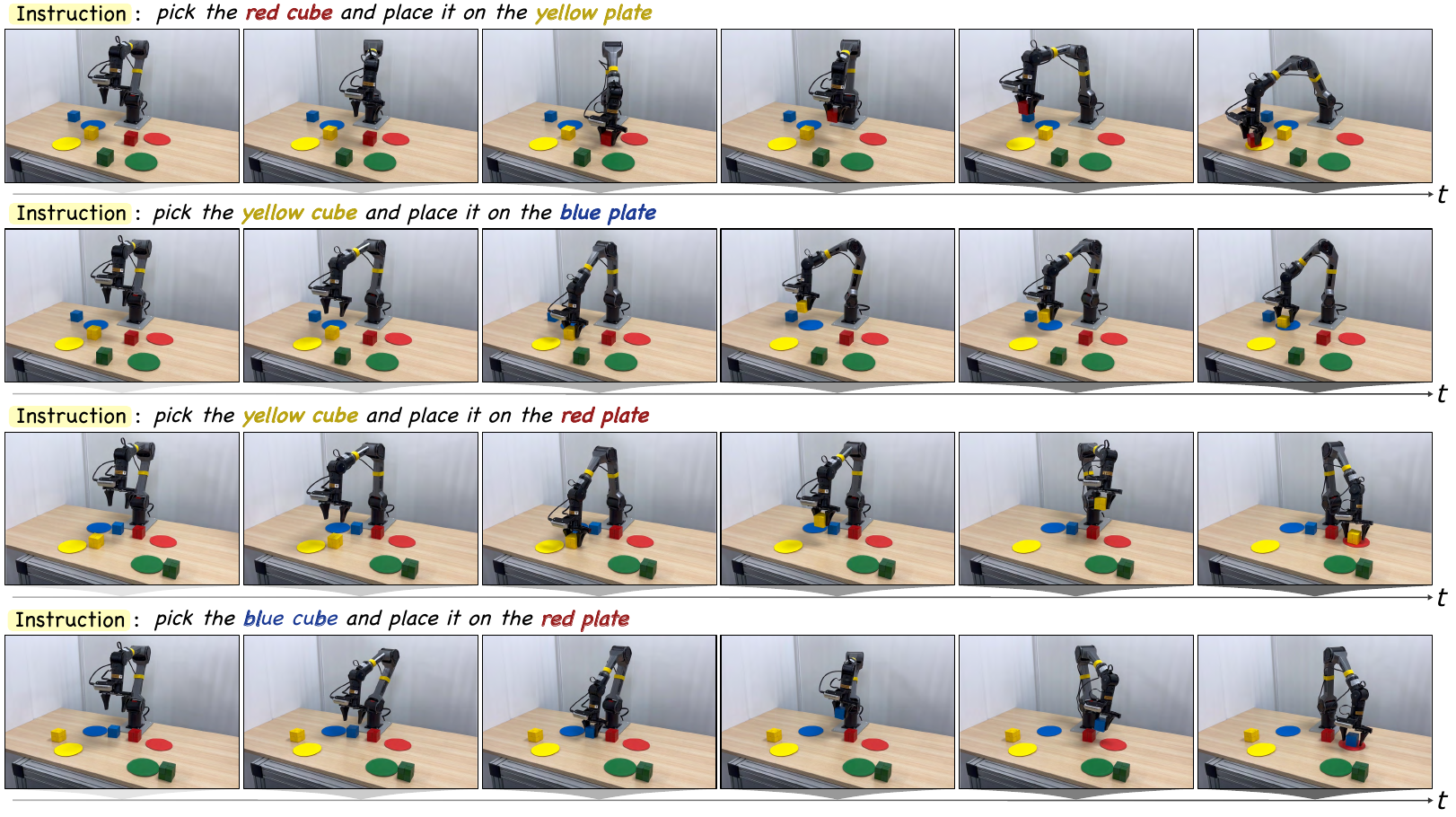}
\caption{
Additional real-robot rollouts of CRAFT on undemonstrated \textsc{Pick-Place} combinations.
Each row shows a different instruction and its execution over time.
CRAFT picks the instructed cube and places it on the instructed plate, although the corresponding cube--plate combination is absent from the demonstrations.
}
\label{fig:real-additional}
\end{figure}

Figure~\ref{fig:real-additional} shows additional real-robot rollouts of CRAFT on undemonstrated \textsc{Pick-Place} combinations.
Across these examples, CRAFT picks the cube specified by the instruction and places it on the specified plate, despite the corresponding cube--plate combinations being absent from the demonstrations.

\end{document}